\documentclass{article}

\PassOptionsToPackage{numbers, compress}{natbib}
 \usepackage[main, final]{neurips_2026}
\usepackage[utf8]{inputenc} % allow utf-8 input
\usepackage[T1]{fontenc}    % use 8-bit T1 fonts
\usepackage{hyperref}       % hyperlinks
\usepackage{url}            % simple URL typesetting
\usepackage{booktabs}       % professional-quality tables
\usepackage{amsfonts}       % blackboard math symbols
\usepackage{nicefrac}       % compact symbols for 1/2, etc.
\usepackage{microtype}      % microtypography
\usepackage[table]{xcolor}  % colors (table option for \rowcolor)

\definecolor{impgreen}{rgb}{0.0, 0.5, 0.0}
\definecolor{degred}{rgb}{0.7, 0.0, 0.0}
\newcommand{\imp}[1]{\rlap{\,\textsubscript{\textcolor{impgreen}{#1}}}}
\newcommand{\degr}[1]{\rlap{\,\textsubscript{\textcolor{degred}{#1}}}}

\usepackage{amsmath}
\usepackage{amssymb}
\usepackage{mathtools}
\usepackage{amsthm}

\usepackage{etoc}
\usepackage[capitalize,noabbrev]{cleveref}
\usepackage{caption}        % \captionof for cross-type captions
\usepackage{wrapfig}        % wrapfigure for text-wrapped floats

\theoremstyle{plain}
\newtheorem{theorem}{Theorem}[section]

\theoremstyle{definition}
\newtheorem{definition}[theorem]{Definition}

\theoremstyle{remark}

\usepackage{booktabs}
\usepackage{siunitx}
\usepackage{amssymb}
\usepackage{amsmath}
\usepackage{multirow}
\usepackage{graphicx}
\usepackage{amssymb}
\usepackage{pifont}
\usepackage{tabularx}
\newcommand{\name}{SAGE}
\newcommand{\eg}{\emph{e.g.}}

\title{Scalable Adaptation of 3D Geometric Foundation Models via Weak Supervision from Internet Video}

\author{%
  Zihui Gao$^{2,1}$ \quad Ke Liu$^{3}$ \quad Donny Y.~Chen$^{4}$ \quad Duochao Shi$^{2}$ \quad Guosheng Lin$^{1}$ \\
  \bfseries Jia-Wang Bian$^{1\dagger}$ \quad Hao Chen$^{2\dagger}$ \quad Chunhua Shen$^{2,5}$ \\[6pt]
  \normalfont\small $^{1}$Nanyang Technological University, Singapore \quad
  $^{2}$Zhejiang University, State Key Lab of CAD \& CG \\
  \normalfont\small $^{3}$ Zhejiang University, School of Software Engineering \quad
  $^{4}$Monash University \quad $^5$Ant Group
}

\begin{document}

\maketitle
{\renewcommand{\thefootnote}{}\footnotetext{$^{\dagger}$Corresponding authors: Hao Chen (\texttt{haochen.cad@zju.edu.cn}) and Jia-Wang Bian (\texttt{jiawang.bian@ntu.edu.sg}).}}

% \begin{abstract}
%   The abstract paragraph should be indented \nicefrac{1}{2}~inch (3~picas) on both the left- and right-hand margins. Use 10~point type, with a vertical spacing (leading) of 11~points. The word \textbf{Abstract} must be centered, bold, and in point size 12. Two line spaces precede the abstract. The abstract must be limited to one paragraph.
% \end{abstract}

\begin{abstract}
Geometric foundation models show promise in 3D reconstruction. However, their progress is severely constrained by the scarcity of diverse, large-scale 3D annotations. 
% internet video用起来存在问题
% While Internet videos offer virtually unlimited raw data, utilizing them for training is hindered by the lack of ground-truth geometry and the presence of significant noise. 
%While unconstrained Internet videos offer virtually unlimited raw data, utilizing them for geometric learning is challenging due to the absence of ground-truth geometry and the presence of observational noise.
While static-scene Internet videos are abundant and diverse, utilizing them as a scaling source for geometric learning is challenging due to the absence of ground-truth geometry and the presence of observational noise.
% While Internet videos offer a virtually unlimited source of natural scene data, leveraging them for training remains an open challenge due to the lack of ground-truth geometry and the presence of significant noise. 
% 我们去解决用internet video的问题
% To extract robust learning signals from unpsed and unlabelled videos, we introduce a hybrid supervision mechanism: we leverage sparse Struture-from-Motion (SfM) points as geometric anchors to prevent structural collapse, while enforcing dense spatiotemporal consistency via differentiable novel view synthesis to learn fine-grained details. 
%Furthermore, we introduce a regularization strategy using a small fraction of anchor data to migate
% To address this, we propose \name, a framework for the scalable adaptation of pretrained geometric models using weakly-supervised Internet video. 
% 方法
% We introduce a hybrid supervision mechanism that distills spatiotemporal consistency into robust geometric signals. Specifically, our approach utilizes sparse Structure-from-Motion points as geometric anchors to prevent structural collapse, while enforcing dense photometric consistency via differentiable 3D Gaussian rendering to learn fine-grained surface details. Furthermore, we employ a regularization strategy using anchor data to mitigate catastrophic forgetting during adaptation.
%
To address this, we propose \textit{SAGE}, a framework for \textit{S}calable \textit{A}daptation of \textit{GE}ometric foundation models from raw video streams. 
SAGE leverages a hierarchical mining pipeline to transform videos into training trajectories and hybrid supervisions: (1) Informative training trajectory selection; %via spatiotemporal consistency;% Trajectory to select informative samples; 
(2) Sparse Geometric Anchoring via SfM point clouds for global structural guidance; and (3) Dense Differentiable Consistency via 3D Gaussian rendering for multi-view constraints. 
Besides, we also introduce a regularization strategy that replays a small fraction of supervised data to prevent catastrophic forgetting.
% To handle the sparsity and noise inherent in SfM outputs, we design a hybrid supervision mechanism with three key components: sparse SfM point clouds serve as geometric anchors 
% to prevent structural collapse; dense photometric consistency enforced via 
% differentiable 3D Gaussian rendering compensates for sparse coverage and 
% provides multi-view constraints; and a regularization strategy using anchor
% data to prevent catastrophic forgetting. 
%during adaptation.
%效果
% Extensive experiments show that \name~significantly enhances zero-shot generalization, reducing Chamfer Distance by 20-42\% on unseen benchmarks (7Scenes, TUM-RGBD, Matterport3D) compared to state-of-the-art baselines. 
% To our knowledge, \name~pioneers the adaptation of geometric foundation models via Internet video, establishing a scalable paradigm for general-purpose 3D learning.
% We validate \name~on \textbf{two architecturally distinct backbones} — the CroCo-based MV-DUSt3R and the ViT-based VGGT — and observe consistent improvements on both, indicating that the recipe is architecture-agnostic rather than tied to a specific pretraining paradigm. On unseen benchmarks (7Scenes, TUM-RGBD, Matterport3D), \name~reduces Chamfer Distance by 20--42\% over the corresponding pretrained backbone, \textbf{without using any new 3D annotations}.
% \name~establishes Internet video as a scalable adaptation source for geometric foundation models — orthogonal to, and complementary with, the recent line of work that scales \emph{architecture and 3D-supervised data}.
Applied to two architecturally distinct backbones, \name~reduces Chamfer Distance on unseen benchmarks (7Scenes, TUM-RGBD, Matterport3D) by 20--42\% with MV-DUSt3R and consistently improves VGGT. \name~establishes Internet video as a viable and scalable adaptation source for geometric foundation models.
Code is available at \url{https://github.com/aim-uofa/SAGE}.
%, complementary to the line of work that scales architecture and high-cost 3D supervision.
% Extensive experiments on two distinct backbones (MV-DUSt3R and VGGT) show that \name~consistently reduces Chamfer Distance by 20--42\% on unseen benchmarks (7Scenes, TUM-RGBD, Matterport3D). 
\end{abstract}

\section{Introduction}
\label{sec:intro}

\begin{figure}[ht]
\centering
\includegraphics[width=.99 \textwidth]{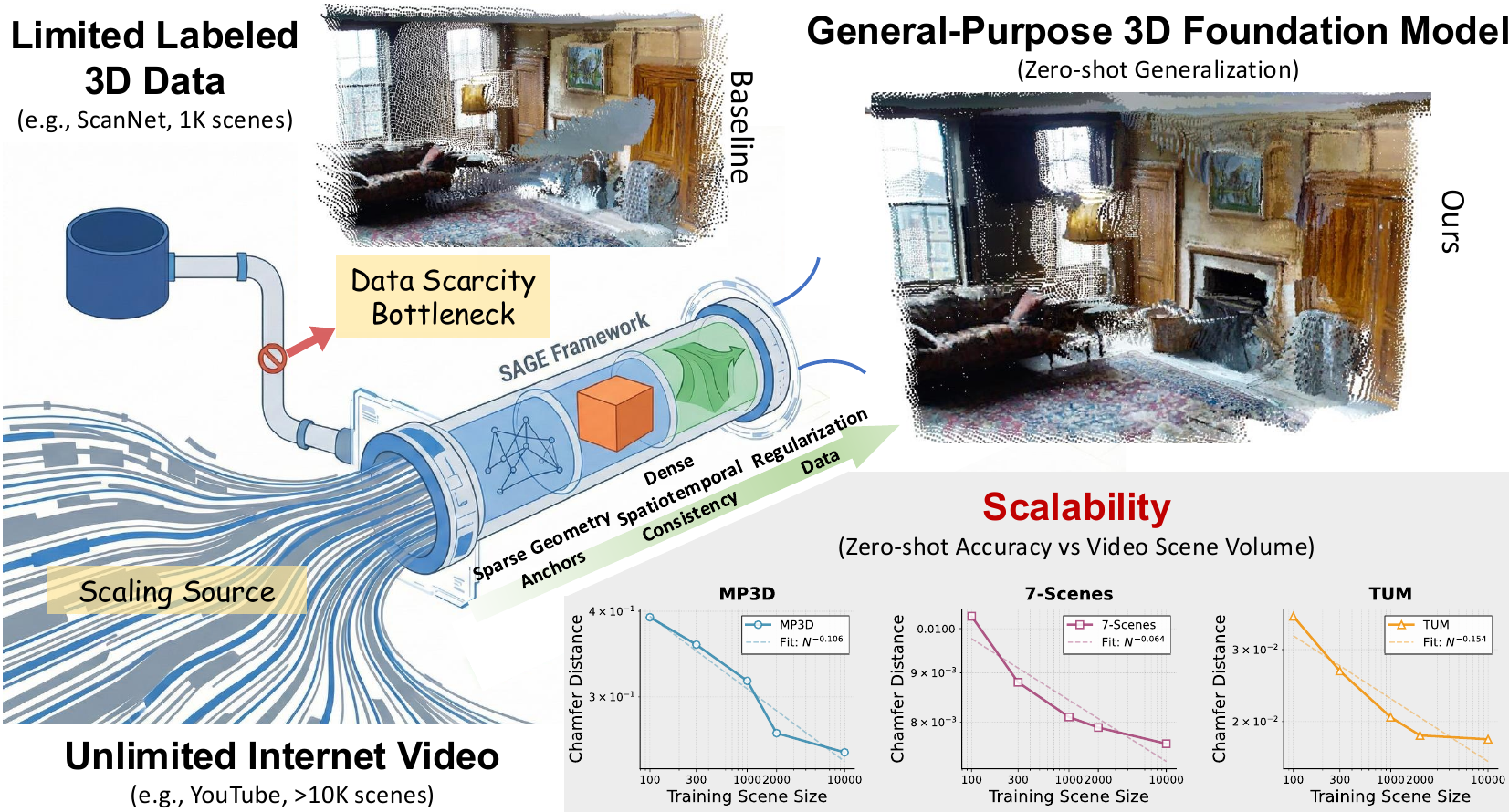}
\caption{Overview of the SAGE framework for scaling a General-Purpose 3D Foundation Model. To overcome the Data Scarcity Bottleneck inherent in limited labeled 3D datasets, we propose a pipeline that leverages abundant static-scene Internet videos to achieve robust zero-shot generalization on complex unseen scenes. Bottom right: Scalability analysis demonstrates zero-shot reconstruction performance as training data scales from 100 to 10K video scenes, showing consistent improvement with increased data volume across various benchmarks (MP3D, 7-Scenes, and TUM).}
\label{fig:final_teaser}
\vspace{-5pt}
\end{figure}

% \begin{figure}[ht]
% \centering
% \includegraphics[width=0.5\textwidth]{sec/figure/scaling_10k.pdf}
% \caption{Zero-shot reconstruction performance as training data scales from 100 to 10K 
% video scenes, showing consistent improvement with increased data volume. 
% %Each scene yields at least tens of training samples from video frames.
% % Top: Standard training framework of existing 3D geometric foundation models. Bottom: 
% \textit{\textbf{10$\times$} increase in scene diversity} compared to standard 3D datasets (like ScanNet).
% }
% \label{fig:scaling}
% \end{figure}

% \begin{figure}[ht]
% \centering
% \includegraphics[width=0.5\textwidth]{sec/figure/teaser_ours.pdf}
% \caption{\textbf{Qualitative comparisons of 3D reconstruction.}
% Reconstructed point clouds by MV-DUSt3R, VGGT, and \name, showing more complete and structurally coherent geometry in our results. 
% % (c) Additional relative reconstruction results on diverse datasets. Leveraging video data enables the model to observe richer scene variations, resulting in improved reconstruction quality.
% }
% \label{fig:teaser}
% \end{figure}
% fm很强，但是受限于监督数据少（昂贵）
Recent advances in 3D geometric foundation models, such as DUSt3R~\cite{wang2024dust3r} and MASt3R~\cite{leroy2024mast3r}, have reshaped the pipeline of reconstructing 3D geometry from unposed 2D images~\cite{wang2026ffsurvey}. These methods move beyond traditional Structure-from-Motion (SfM)~\cite{snavely2006sfm} pipelines with end-to-end data-driven networks that directly estimate pixel-aligned pointmaps in a unified coordinate system. Further developments like MV-DUSt3R~\cite{tang2024mvdust3r}, Fast3R~\cite{yang2025fast3r}, and VGGT~\cite{wang2025vggt} extend the framework to multi-view inputs, enabling dense 3D reconstruction in a single forward pass. These 3D foundation models achieve impressive results even with extremely sparse inputs or in zero-shot settings, demonstrating strong generalization across unseen scenes.
%% 但是受限于监督数据少（昂贵
However, \textit{their further scaling is severely bottlenecked by the scarcity of high-quality 3D annotations}. Unlike 2D vision models trained on billions of images, 3D models heavily rely on sensor-captured datasets like ScanNet, which are costly to scale and restricted to static, indoor environments. This reliance on limited supervised data results in a significant distribution shift when applying these models to diverse, in-the-wild scenes.

% 互联网数据多，但是无标注，有噪声
%% 互联网数据多
To overcome this data scalability bottleneck, Internet videos emerge as a promising alternative, offering abundant coverage of diverse real-world static scenes.
For example, the RealEstate10K~\cite{zhou2018re10k} dataset, sourced from YouTube videos, contains over \textbf{10,000} unique scenes, far more than 3D datasets like ScanNet~\cite{dai2017scannet}, which has only around \textbf{1,000} depth sensor-annotated scenes. 
%% 直接用伪标签会失败
However, \textit{unlocking this potential is non-trivial due to the lack of ground-truth geometry and camera poses.} 
A straightforward approach, which we refer to as Video Depth Supervision (VDS), fine-tunes models using pseudo-depth labels from video depth estimators~\cite{yang2024depthanyvideo}, but often fails to improve or even degrades performance 
%\deleted{due to temporal inconsistencies and accumulated noise}, 
in our empirical analysis.
The main reason is that the predicted video depths contain substantial noise and often lack multi-view consistency, making accurate alignment difficult and ultimately degrading reconstruction quality.
%% 引出问题，如何用
Therefore, the core challenge lies in extracting robust supervision signals from noisy, unlabelled video streams without drifting into degenerate solutions.

% 讲Vi3D方法
To tackle this challenge, we propose \name, a framework for the \textit{S}calable \textit{A}daptation of \textit{GE}ometric foundation models. 
Specifically, we develop a spatio-temporal mining pipeline to curate informative training trajectories as well as complementary supervisions from raw videos. We first perform stochastic trajectory sampling to curate frame sequences balancing viewpoint diversity and geometric overlap, providing the data foundation for multi-view learning. Upon these curated trajectories, we utilize sparse point clouds derived from SfM as geometric anchors to provide reliable global structural guidance. To complement these sparse anchors, we leverage differentiable 3D Gaussian Splatting to enforce dense photometric consistency across video frames. Furthermore, to stabilize the adaptation process on diverse video distributions, we introduce a regularization strategy using a small fraction of supervised replay data, effectively mitigating catastrophic forgetting of the original geometric priors.

% Scalability，generalization，contribution
Extensive experiments demonstrate that \name~turns Internet video into an effective and scalable supervision source. 
As shown in Figure~\ref{fig:final_teaser}, zero-shot accuracy improves consistently as the amount of video data scales from 100 to 10K scenes, following a clear log-linear trend without saturation. 
% This confirms that our weak supervision strategy successfully converts virtually unbounded raw video into effective and scalable training signals, without saturation.
We validate our approach by adapting pretrained multi-view reconstruction models 
% (MV-DUSt3R~\cite{tang2024mvdust3r}) 
using over 10,000 video clips from RealEstate10K and DL3DV~\cite{ling2024dl3dv}, which is 10 times larger than the standard 3D datasets.
Both in-distribution performance and zero-shot generalization capabilities are improved by our \name.
On unseen benchmarks (7Scenes, TUM-RGBD, Matterport3D), our method reduces the Chamfer Distance of MV-DUSt3R by 20--42\% and consistently improves VGGT. 

Importantly, \name~provides an orthogonal scaling axis to recent efforts that improve 3D foundation models through larger architectures and larger curated 3D corpora, such as VGGT~\cite{wang2025vggt}, $\pi^3$~\cite{wang2025pi}, and Depth Anything 3~\cite{lin2025da3}.
Rather than training a larger model from scratch with more supervised 3D data, \name~adapts existing pretrained backbones using cheap and scalable Internet video. 
To show that this recipe is not tied to a specific architecture, we instantiate \name~on two representative backbones with different multi-view processing paradigms: MV-DUSt3R, which uses an encoder--decoder architecture with cross-view cross-attention and predicts pointmaps in a selected reference frame; and VGGT, which uses a single transformer with alternating frame-wise and global self-attention, together with multi-task heads for point maps, depth, and camera parameters. 
Consistent improvements on both backbones suggest that video-based adaptation is complementary to architectural and supervised-data scaling.

Our main contributions are summarized as follows:
% (1) We propose \name, the first framework to enable the scalable adaptation of feed-forward geometric foundation models using unlabelled Internet video.\\
% (2) We design a robust hybrid supervision strategy that integrates sparse geometric anchors, dense spatiotemporal consistency, and regularization to effectively learn from noisy video data.\\
% (3) We instantiate \name~on two representative backbones, MV-DUSt3R and VGGT, and obtain consistent gains across four benchmarks, establishing Internet video as a viable scaling source for general-purpose 3D foundation models.

\begin{itemize}
    \item We propose \name, a framework for scalable adaptation of pretrained feed-forward geometric foundation models using unlabelled static-scene Internet video, which converts sparse SfM anchors and frozen-renderer multi-view consistency into task-preserving supervision.
    \item We design a robust hybrid supervision strategy that integrates sparse geometric anchors, dense spatiotemporal consistency, and regularization to effectively learn from noisy video data.
    \item We instantiate \name~on two representative backbones, MV-DUSt3R and VGGT, and obtain consistent gains across four benchmarks, establishing Internet video as a viable scaling source for general-purpose 3D foundation models.
\end{itemize}

% We demonstrate significant improvements in zero-shot generalization across 4 benchmarks and the scalability of our approach, establishing Internet video as a viable resource for evolving general-purpose 3D models.

\section{Related work}
\label{sec:related_work}

\subsection{Geometric Foundation Models \& Data Scarcity}
Traditional approaches to pose-free multi-view 3D reconstruction include Structure-from-Motion (SfM)~\cite{snavely2006sfm,crandall2012sfm,schonberger2016sfm,cui2017hsfm} and Multi-View Stereo (MVS)~\cite{mvs1,yao2018mvsnet,goesele2006mvs3}. While effective, they rely on complex multi-stage optimization.
Recently, the field has shifted towards Geometric Foundation Models such as DUSt3R~\cite{wang2024dust3r} and VGGSfM~\cite{wang2024vggsfm}, which jointly estimate camera parameters and 3D structure in a data-driven manner. Extensions including MV-DUSt3R, Fast3R~\cite{yang2025fast3r}, VGGT~\cite{wang2025vggt} and $\pi^3$~\cite{wang2025pi} further scale this to multi-view inputs.

\textit{However, these models are severely bottlenecked by the scarcity of 3D supervision.} Unlike 2D models trained on web-scale data, they rely on costly annotated datasets like ScanNet, limiting their generalization to in-the-wild distributions. While methods like CUT3R~\cite{wang2025cut3r} use video sequences, they %\deleted{rely on provided camera parameters}. 
%\added{directly supervises on noisy SfM poses. In contrast, we formulate this as a weak supervision problem and employ a hybrid objective to improve robustness.}
directly supervise on noisy SfM poses. 
% In contrast, we formulate this as a weak supervision problem and employ a hybrid objective to improve robustness. 
In contrast, our approach addresses this scalability bottleneck by leveraging unlabelled Internet video. We propose a weakly-supervised adaptation framework that combines sparse geometric anchors with multi-view consistency, enabling foundation models to learn from diverse, real-world data without ground-truth poses.
Concurrently, VGGT-$\Omega$~\cite{wang2026vggtomega} scales foundation-model training to large video corpora using dense annotations from COLMAP bundle adjustment and MVS, together with teacher--student self-training.
\name~is complementary: rather than re-training a foundation model at scale, it adapts released checkpoints through lightweight post-training, using only sparse SfM points and rendering consistency.

\subsection{Differentiable Rendering as Weak Supervision}

Recent works PixelSplat~\cite{charatan2024pixelsplat}, MVSplat~\cite{chen2024mvsplat}, and DepthSplat~\cite{xu2024depthsplat} predict 3D Gaussian representations from sparse views for novel view synthesis~\cite{wang2025zpressor,wang2026volsplat,wang2026trisplat}. Compared with optimization-based methods like 3DGS~\cite{3dgs,gao2025surfacesplat}, these feedforward approaches offer better generalization. Extensions such as MVSplat360~\cite{chen2024mvsplat360} and LatentSplat~\cite{wewer2024latentsplat} further improve rendering flexibility.

\textit{While these methods target view synthesis, we repurpose the differentiable nature of 3D Gaussians as a dense supervision mechanism.}
In our framework, the Gaussian renderer serves as a differentiable loss function that enforces photometric consistency across video frames. This allows us to propagate dense gradients to the underlying geometry in regions undefined by sparse points. Unlike MV-DUSt3R, which trains a Gaussian head for downstream rendering, we utilize this differentiable consistency to robustly adapt the geometric backbone itself.

\subsection{Scalable Representation Learning from Video}

Self-supervised learning on video sequences leverages spatiotemporal consistency as an intrinsic supervision signal. Early works~\cite{zhou2017unsupervised} use image-warping losses, while subsequent methods~\cite{bian2019unsupervised,godard2019digging} introduce scale-consistent predictions for depth estimation. 
More recently, 
CroCo~\cite{weinzaepfel2022croco,weinzaepfel2023croco} extends masked image modeling to cross-view settings for representation learning. 
E-RayZer~\cite{zhao2025rayzer} further advances this line by learning 
geometrically grounded representations through self-supervised 3D Gaussian 
reconstruction from unlabeled video.
NAS3R~\cite{huang2026nas3r} similarly trains, or self-supervisedly adapts, multi-view geometric models such as MASt3R and VGGT with differentiable Gaussian rendering.
These methods primarily target novel-view synthesis and camera estimation, with photometric rendering as the main objective.
In contrast, \name~targets surface-geometry reconstruction evaluated with point-cloud metrics: SfM anchors provide the primary geometric supervision, while rendering through a \textit{frozen} Gaussian head serves only as an auxiliary dense constraint, forcing gradients to refine the backbone's geometry rather than the renderer.

\textit{While most of these methods train task-specific networks from scratch, they share the principle of deriving supervision from video consistency. }
Our work extends this principle to the adaptation of pre-trained geometric foundation models. By integrating sparse geometric constraints (bias-reduction) with dense video consistency (variance-reduction), we propose a hybrid learning framework that effectively fine-tunes general-purpose models on noisy, large-scale video data, significantly enhancing their zero-shot generalization.
\section{Method}
\label{sec:method}

\begin{figure*}
\centering
\includegraphics[width = 0.95\textwidth]{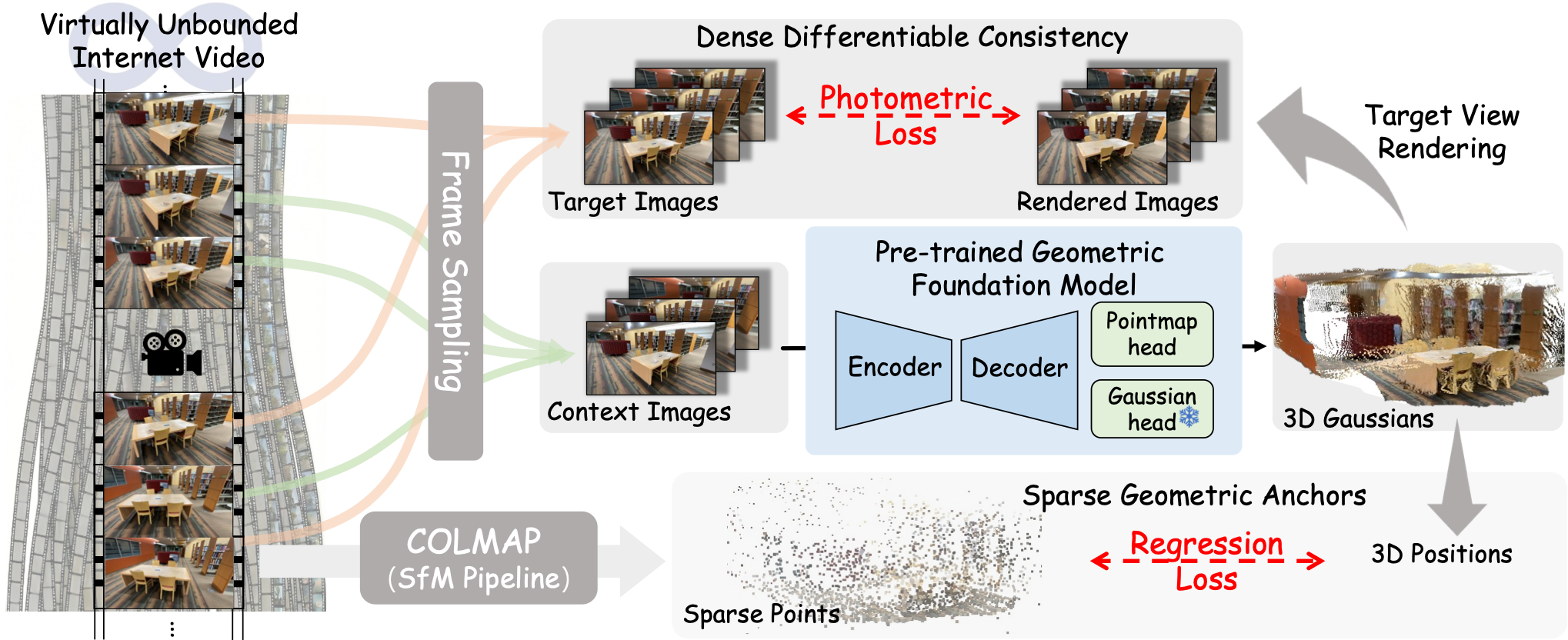}
  \caption{Illustration of \name. For each video sequence, we sample context frames as model inputs and designate target frames for novel-view supervision, providing photometric constraints to refine the reconstructed 3D point cloud. Furthermore, we incorporate sparse 3D point clouds that provide consistent geometric constraints, complementing the photometric supervision from the target views.}
  \label{fig:method}
  \vspace{-10.5pt}
\end{figure*}

We address the problem of scaling up a pre-trained geometric foundation model, $f_\theta$, by adapting it to the domain of unlabelled Internet videos. We formulate this as a weakly-supervised optimization problem, where the goal is to leverage intrinsic spatiotemporal consistency to refine the model without requiring ground-truth 3D annotations.

\subsection{Preliminary \& Problem Formulation}

We formulate the training of geometric foundation models as a scalable optimization problem, transitioning from a data-scarce supervised regime to a
data-abundant weakly-supervised regime.

\subsubsection{Task Instantiation} 

We focus on \textit{unconstrained sparse-view 3D reconstruction}.
\begin{definition}[pose-free sparse-view 3D reconstruction]
   Let $f_\theta: \mathcal{I}^N \to \mathcal{X}$ denote a geometric foundation model parameterized by $\theta$. Given a set of unposed RGB images $I \in \mathcal{I}^N$, the model directly predicts a dense 3D representation $\mathcal{X}$ (specifically, pixel-aligned 3D pointmaps) without requiring ground-truth camera poses. 
\end{definition}
% Our objective is to leverage the unbounded scale of Internet video to transcend the generalization limits of $f_\theta$.
Our objective is to leverage the abundance of static-scene Internet video to push the generalization limits of $f_\theta$.

\subsubsection{Bottleneck \& Scaling Source}
\paragraph{Supervised Regime (Bottleneck)} Existing models are typically initialized on a supervised dataset $\mathcal{D}_{sup} = \{(I, Y)\}$, where $Y$ represents expensive GT 3D annotations (e.g., depth, poses). 
While $\mathcal{D}_{sup}$ offers high-fidelity supervision, it is fundamentally non-scalable due to the prohibitive cost of 3D data acquisition. Consequently, $|\mathcal{D}_{sup}|$ remains small, restricting the model's exposure to a narrow distribution of %\deleted{indoor} 
scenes and limiting its zero-shot generalization capabilities 
%\added{to in-the-wild scenarios}
to in-the-wild scenarios.
\paragraph{Weakly-Supervised Regime (Scaling Source)} To address this bottleneck, we introduce a large collection of static-scene Internet videos, denoted as $\mathcal{D}_{video} = \{V_j\}$.
In contrast to $\mathcal{D}_{sup}$, this dataset is scalable ($|\mathcal{D}_{video}| \gg |\mathcal{D}_{sup}|$) but \textit{unlabelled}: it lacks explicit 3D geometry ($Y=\emptyset$) and camera poses. However, it contains rich intrinsic supervision encoded in physical constraints, specifically spatiotemporal consistency.

\subsubsection{Optimization Objective} 
We aim to learn an optimal parameter set $\theta^*$ that absorbs geometric knowledge from massive video data while preserving the structural priors learned from the supervised regime. We formulate it as a constrained optimization problem:
\begin{align}
    \min_{\theta} &\underbrace{\mathbb{E}_{V \sim \mathcal{D}_{video}} [\mathcal{L}_{consistency}(f_\theta(V))]}_{\text{Scaling via Weak Supervision}} + ~\gamma \cdot \underbrace{\mathbb{E}_{(I, Y) \sim \mathcal{D}_{sup}} [\mathcal{L}_{sup}(f_\theta(I), Y)]}_{\text{Foundation Regularization}},
    \label{eq:loss_all}
\end{align}
where $\mathcal{L}_{consistency}$ represents a hybrid loss function designed to distill robust geometric signals from noisy video consistency, and the second term serves as a regularization constraint to prevent catastrophic forgetting.

\subsection{Scaling with Spatio-Temporal Weak Supervision}
% To facilitate the scaling of 3D foundation models through sparse-view 3D reconstruction, central is \textbf{how to curate informative training samples and latent supervision from massive, in-the-wild video corpora}.
A fundamental requirement for scaling 3D foundation models is the \textbf{ability to harvest informative training trajectories and latent supervision from massive, unconstrained video streams}.
% Motivated by this, we first introduce a video data processing pipeline in App.~\ref{app:data_pocess}, followed by a spatio-temporal mining pipeline that extracts informative training trajectories from raw videos, as shown in Figure~\ref{fig:method}.
Unlike structured 3D datasets, raw videos are inherently redundant and exhibit inconsistent viewpoint transitions. To address this, we first introduce a video data processing pipeline in App.~\ref{app:data_pocess}, followed by a spatio-temporal mining pipeline that extracts informative training trajectories from raw videos, as shown in Figure~\ref{fig:method}.
First, we curate training trajectories by balancing viewpoint diversity and geometric overlap to ensure representative scene coverage. For these trajectories, we derive two levels of weak supervision: (1) Sparse Geometric Anchors from COLMAP point clouds to provide global structural bias; 
(2) Dense Differentiable Consistency via 3D Gaussian rendering to propagate gradients through video spatiotemporal consistency. This synergy allows our model to effectively optimize Eq.~\ref{eq:loss_all}, providing essential supervision for learning robust 3D geometry from diverse real-world scenes in a scalable and stable manner.

\subsubsection{Spatiotemporal Sampling from Videos}

Unlike prior works~\cite{wang2024dust3r,tang2024mvdust3r} that rely on GT depth to determine image overlap, our approach tackles the \textit{scaling} challenge where only raw video streams are available. We exploit the inherent spatiotemporal continuity of videos as a geometric overlap proxy to mine multi-view training samples.

Given a video sequence, we construct training trajectories $\mathcal{V}$ by balancing viewpoint overlap (ensuring stable supervision) and baseline diversity (promoting geometric learning). While uniform temporal sampling with interval $\Delta t$: 
% \begin{equation}
% \mathcal{V}_{\text{uniform}} = \{ I_{t_0 + i\Delta t} \mid i = 0, 1, \dots, n \}
% \end{equation}
$\mathcal{V}_{\text{uniform}} = \{ I_{t_0 + i\Delta t} \mid i = 0, 1, \dots, n \}$
maintains consistent visual transitions, it constrains viewpoint diversity and introduce motion pattern bias. 

We instead employ \textit{stochastic temporal perturbations} by sampling offsets $\delta_i \sim \mathcal{U}(-\epsilon, \epsilon)$:
% \begin{equation}
% \mathcal{V}_{\text{perturbed}} = \{ I_{t_0 + i\Delta t + \delta_i} \mid i = 0, 1, \dots, n \},
% \end{equation}
$\mathcal{V}_{\text{perturbed}} = \{ I_{t_0 + i\Delta t + \delta_i} \mid i = 0, 1, \dots, n \}$,
which preserves temporal structure for geometric consistency while maximizing camera pose diversity. This curation establishes the data foundation for scalable learning, from which we subsequently derive hierarchical supervision signals (Sec.~\ref{sec:sparse}, \ref{sec:dense}).

%\paragraph{Sparse 3D Supervision}
\subsubsection{Sparse Geometric Anchors}
\label{sec:sparse}
While dense pseudo-depth maps can be easily generated by off-the-shelf models, they often suffer from multi-view inconsistency and scale drift, which can destabilize the scaling process. In contrast, sparse point clouds derived from SfM provide globally consistent geometric anchors, offering a reliable structural bias. Specifically, the sparse point cloud is projected onto each image to generate corresponding sparse depth maps, using the camera intrinsics $\mathbf{K}$ and extrinsics $[\mathbf{R} \,|\, \mathbf{t}]$ from COLMAP. Each 3D point $\mathbf{P} = (x, y, z)$ is projected to image coordinates $(u, v, d)$ via $\begin{bmatrix}u & v & 1\end{bmatrix}^\top \!\sim\! \mathbf{K}[\mathbf{R} \,|\, \mathbf{t}]\begin{bmatrix}x & y & z & 1\end{bmatrix}^\top$, where $d$ denotes the projected depth. Then following DUSt3R and MV-DUSt3R, the 
confidence-aware 3D loss is defined as:
% \begin{equation}
% \mathcal{L}_{\mathrm{conf}} = \sum_{v \in \{1,2\}} \sum_{i \in \mathcal{D}^v} C_i^{v, 1} \cdot \ell_{\mathrm{regr}}(v, i) - \alpha \log C_i^{v, 1}
% \end{equation}
% \begin{equation}
% \ell_{\mathrm{regr}}(v, i) = \left\| \frac{1}{z} X_i^{v,1} - \frac{1}{\bar{z}} \bar{X}_i^{v,1} \right\|
% \end{equation}
\begin{equation}
\mathcal{L}_{\mathrm{anchor}} = \sum_{v \in \{1,...,N\}} \sum_{i \in \mathcal{D}^v} C_i^{v, 1}  \ell_{\mathrm{regr}}(v, i) - \alpha \log C_i^{v, 1}, \ell_{\mathrm{regr}}(v, i) = \left\| \frac{1}{z} X_i^{v,1} - \frac{1}{\bar{z}} \bar{X}_i^{v,1} \right\|,
\end{equation}
% \begin{equation}

% \end{equation}
%$\mathcal{D}^v$ denotes the set of valid supervision pixels in view $v$, and 
where $z$, $\bar{z}$ are scale normalization factors and
$C_i^{v, 1}$ is the predicted confidence score for pixel $i$. The term $X_i^{v, 1}$ represents the predicted 3D coordinate in the reference coordinate. 
This term anchors the predicted geometry to the global SfM structure, mitigating potential geometric drift.
%The normalization term is $z=\operatorname{norm}\left(X^{1,1},..., X^{N,1}\right)$. 

% \paragraph{Multi-View Consistency via Rendering}
%\paragraph{Novel View Supervision from Video Consistency}

\subsubsection{Dense Differentiable Consistency}
\label{sec:dense}
%However, sparse geometric constraints alone are often insufficient, leaving large portions of the point cloud unconstrained and limiting reconstruction accuracy.
Sparse anchors alone leave the majority of the surface undefined. To propagate geometric supervision to dense regions, we derive the Spatiotemporal
Consistency from video. 
%Therefore, we propose a novel view supervision, which takes advantage of the abundance of novel viewpoints in videos.
While prior works optimize 3D Gaussians as a scene representation for 
view synthesis, \textit{we repurpose the splatting pipeline solely as a 
differentiable supervision module}, keeping all Gaussian parameters frozen.
The Gaussian centers are the backbone's predicted 3D points, while opacity, scale, rotation, and color are regressed by a Gaussian head. For MV-DUSt3R we use its pretrained Gaussian head; for point-map backbones without one (\eg, VGGT), we attach a lightweight Gaussian head, train it for 2--3 epochs, and then freeze it during adaptation.

Given a sampled video clip of $M$ temporally ordered frames $\mathcal{V}_{\text{clip}} = \{ I_{t_1}, I_{t_2}, \dots, I_{t_M} \}$, we divide them into \textit{context frames} as inputs and \textit{target frames} as supervision for novel view synthesis. Target views are unseen frames from the same video, used to enforce multi-view consistency during training.
% Specifically, we denote:
% \[
% \mathcal{V}_{\text{clip}} = \{ I_{t_1}, I_{t_2}, \dots, I_{t_M} \}
% \]
To ensure effective supervision, we random sample intermediate frames $t_j \in  \{t_2, t_3, \dots, t_{M-1}\}$ as target views, avoiding extrapolated viewpoints that may introduce artifacts in Gaussian splatting reconstructions.
$\hat{I}_k$ is rendered from differentiable Gaussian Splatting from the predicted point cloud. The loss function is:
\begin{equation}
\mathcal{L}_{\text{ren}} = \sum_{I_k \in \{I_{con}, I_{tgt}\}} \gamma \cdot \operatorname{LPIPS}(I_k, \hat{I}_k) + \left\| I_k - \hat{I}_k \right\|_2,
\end{equation}
where $I_{con}$ and $I_{tgt}$ are context and target frames. This loss enforces image consistency through novel view synthesis of 3D Gaussians, thereby constraining the Gaussian centers and improving the accuracy of the predicted 3D point cloud.

% \subsubsection{Stochastic Spatiotemporal Sampling} %\added{scalability}}

%\paragraph{Geometry-Preserving Regularization}
\subsection{Regularization against Catastrophic Forgetting}

A critical challenge in adapting foundation models to noisy video data is distribution drift, where the model overfits to the video domain and loses its original general-purpose capabilities.
To explicitly mitigate this, we implement the regularization term in Eq.~\ref{eq:loss_all} by mixing a small set of supervised replay data $\mathcal{D}_{sup}$ from the original pre-training distribution.

We define the sampling ratio between video data and supervised replay data as $\eta$. Empirically, we find that a small fraction ($\eta \approx 3\%$) is sufficient to stabilize the optimization landscape. This regularization constrains the model parameters to remain in the valid manifold of plausible 3D geometries, preventing catastrophic forgetting while allowing the model to absorb new diversity from Internet videos.

% \subsection{Frames Sampling from Video Sequences}
% Given unlabeled video sequences, we first apply COLMAP preprocessing to recover camera poses and sparse point clouds. Based on the estimated poses, we explore two strategies for selecting multi-view input frames: uniform sampling with temporal perturbation and pose-aware trajectory sampling. These sampling approaches enable the model to observe diverse scene configurations while ensuring sufficient viewpoint variation for supervision.

%%% 这里如果没有采用uniform的采样，就可以直接说是stochastic sampling了，作为一种数据增强方式，这样也能解决cvpr reviewer的问题。

\section{Experiments}
\label{sec:experiment}
% 实验结果
%% 主要结果（+不同数据量结果）
%% 消融实验 
%%% 三个loss的作用

% Discussion
%%% 如何选择 Anchor data 和 video data的数据比例
%%% 如何选择 数据 （红绿的比例）
%%% 对下游其他任务的影响（pose estimation）

% \subsection{Experimental Setup}

We use the pretrained MV-DUSt3R~\cite{tang2024mvdust3r} and VGGT~\cite{wang2025vggt} as our baseline, originally trained on 3D ground-truth annotated datasets, and we fine-tune them with Internet video data to enhance its 3D reconstruction performance. Video preprocessing is a one-time, cacheable sparse-SfM pass (${\sim}15$\,s per clip; over 13K clips in 5--6 hours on four RTX 4090 GPUs), and adaptation completes in under two days on the same hardware. For more details about dataset, Implementation details and Evaluation metrics, please refer to App.~\ref{app_details}.

\begin{table*}[htbp]
  \centering
  \footnotesize
  \setlength{\tabcolsep}{2pt}
  \renewcommand{\arraystretch}{1.15}
  
  % \caption{
  %   \textbf{8-view reconstruction results} on four benchmarks, all evaluated at $224\times224$ for a fair comparison.
  %   The table is organized along two complementary axes:
  %   \emph{(A)} feed-forward foundation models trained \textbf{from scratch with full 3D supervision} (sensor depth and ground-truth poses) on large curated 3D datasets;
  %   \emph{(B,C)} our \name~\textbf{label-free adaptation recipe} that fine-tunes a pretrained backbone using \textbf{only unlabelled Internet video} (zero new 3D annotations).
  %   The two axes are complementary rather than competing: (A) scales architecture and 3D supervision; (B,C) scales data via cheap unlabelled video.
  %   \name~delivers consistent gains over both backbones it adapts (MV-DUSt3R, VGGT), and on the zero-shot benchmarks 7Scenes and TUM-RGBD it matches or surpasses the strongest 3D-supervised foundation models on \emph{both} DAc and CD.
  %   \textbf{VDS} (Video Depth Supervision) is a prior video-based recipe that fine-tunes MV-DUSt3R on pseudo depth from Depth Anything Video, included as a video-recipe comparison.
  %   \textbf{SAGE ($n$K)} denotes adaptation using $n$K video clips.
  %   Subscripts in \textcolor{impgreen}{green}/\textcolor{degred}{red} show $\Delta$ over the corresponding pretrained backbone.
  %   \underline{Underline}: best within group (A); \textbf{bold}: best within groups (B,C).
  %   $\ast$ denotes global alignment (GA) optimization.
  % }
\caption{
  \textbf{8-view reconstruction results} on four benchmarks at $224\times224$ resolution.
  Group~(A) includes feed-forward foundation models trained from scratch with full 3D supervision, while Groups~(B,C) adapt pretrained backbones using only Internet video, with no additional depth or pose annotations.
  \name~consistently improves both MV-DUSt3R and VGGT, and matches or surpasses the strongest 3D-supervised models on the zero-shot 7Scenes and TUM-RGBD benchmarks.
  \textbf{VDS} denotes video-depth supervision using pseudo depth from Depth Any Video.
  \textbf{SAGE ($n$K)} uses $n$K video clips for adaptation.
  Subscripts in \textcolor{impgreen}{green}/\textcolor{degred}{red} show $\Delta$ over the corresponding pretrained backbone.
  \underline{Underline}: best within Group~(A); \textbf{bold}: best within Groups~(B,C).
  % $\ast$ denotes global alignment optimization.
}

  \label{tab:main_results}

  \begin{tabular}{lcccccccc}
    \toprule
    & \multicolumn{2}{c}{\textbf{In-distribution}}
    & \multicolumn{6}{c}{\textbf{Zero-shot Generalization}} \\
    \cmidrule(lr){2-3} \cmidrule(lr){4-9}
    \multirow{2}{*}{\textbf{Method}}
    & \multicolumn{2}{c}{\textbf{ScanNet}}
    & \multicolumn{2}{c}{\textbf{7Scenes}}
    & \multicolumn{2}{c}{\textbf{TUM-RGBD}}
    & \multicolumn{2}{c}{\textbf{Matterport3D}} \\
    \cmidrule(lr){2-3} \cmidrule(lr){4-5} \cmidrule(lr){6-7} \cmidrule(lr){8-9}
    & DAc@0.2~$\uparrow$ & CD~$\downarrow$
    & DAc@0.2~$\uparrow$ & CD~$\downarrow$
    & DAc@0.2~$\uparrow$ & CD~$\downarrow$
    & DAc@0.5~$\uparrow$ & CD~$\downarrow$ \\
    \midrule
    \multicolumn{9}{l}{\textit{(A) Foundation models trained from scratch with full 3D supervision (depth / poses)}} \\
    \midrule
    Spann3R~\cite{wang2024spann3r}
      & 59.0 & 13.48 & 90.0 & 1.48 & 35.7 & 7.06 & 15.8 & 69.7 \\
    DUSt3R~\cite{wang2024dust3r}
      & 88.0 & 4.50  & 89.5 & 1.50 & 71.4 & 4.65 & 45.4 & 44.1 \\
    MapAnything~\cite{keetha2025mapanything}
      & 79.5 & 3.05  & 88.1 & 1.56 & 82.0 & 2.96 & 51.8 & 24.0 \\
    $\pi^3$~\cite{wang2025pi}
      & \underline{97.0} & \underline{0.95} & \underline{95.0} & \underline{0.80}
      & \underline{92.7} & 1.90 & 57.4 & 22.9 \\
    Depth Anything 3~\cite{lin2025da3}
      & 89.8 & 2.45 & 91.5 & 0.84
      & 85.7 & \underline{1.69} & \underline{67.3} & \underline{17.1} \\
    \midrule
    \multicolumn{9}{l}{\textit{(B) MV-DUSt3R + video-based adaptation (no new 3D annotations)}} \\
    \midrule
    MV-DUSt3R~\cite{tang2024mvdust3r} \textit{(pretrained backbone)}
      & 88.2 & 1.79 & 89.6 & 0.95 & 87.5 & 2.60 & 30.3 & 43.2 \\
    \quad + VDS \textit{\footnotesize (prior recipe)}
      & 90.0\imp{+1.8}     & 1.78\imp{-0.01}
      & 92.5\imp{+2.9}     & 0.89\imp{-0.06}
      & 84.7\degr{-2.8}  & 2.68\degr{+0.08}
      & 14.3\degr{-16.0} & 82.1\degr{+38.9} \\
    \quad + SAGE (1K)
      & 90.0\imp{+1.8}  & \textbf{1.61}\imp{-0.18}
      & 96.0\imp{+6.4}  & 0.79\imp{-0.16}
      & 91.7\imp{+4.2}  & 2.05\imp{-0.55}
      & 47.4\imp{+17.1} & 31.7\imp{-11.5} \\
    \textbf{\quad + SAGE (10K) (Ours)}
      & \textbf{91.2}\imp{+3.0} & 1.69\imp{-0.10}
      & \textbf{98.0}\imp{+8.4} & \textbf{0.76}\imp{-0.19}
      & \textbf{93.1}\imp{+5.6} & \textbf{1.85}\imp{-0.75}
      & 48.0\imp{+17.7}        & 24.9\imp{-18.3} \\
    \midrule
    \multicolumn{9}{l}{\textit{(C) VGGT + video-based adaptation (no new 3D annotations)}} \\
    \midrule
    VGGT~\cite{wang2025vggt} \textit{(pretrained backbone)}
      & 87.0 & 3.30 & 88.5 & 1.13 & 80.4 & 4.24 & 53.1 & 20.7 \\
    \quad + SAGE (10K)
      & 89.5\imp{+2.5} & 2.01\imp{-1.29}
      & 92.0\imp{+3.5} & 0.79\imp{-0.34}
      & 84.3\imp{+3.9} & 2.84\imp{-1.40}
      & \textbf{55.4}\imp{+2.3} & \textbf{18.3}\imp{-2.4} \\
    \bottomrule
  \end{tabular}
\end{table*}

\subsection{Sparse 3D Reconstruction from Multi Inputs}

We evaluate \name~on ScanNet as the in-distribution benchmark, and on 7Scenes, TUM-RGBD, and MP3D to assess zero-shot generalization. 
All methods are compared under the standard 8-view setting at $224\times224$ resolution, with results using 4--24 input views reported in App.~\ref{app:multi-view}. 
Table~\ref{tab:main_results} compares \name~with representative pose-free 3D foundation models, including $\pi^3$~\cite{wang2025pi}, Depth Anything 3~\cite{lin2025da3}, and MapAnything~\cite{keetha2025mapanything}. 
These foundation models in Group~(A) are typically trained from scratch with large-scale, carefully curated 3D supervision, higher-capacity architectures, and/or more involved training recipes. 
In contrast, \name~in Groups~(B,C) adapts pretrained backbones using only unlabelled videos, making the comparison a stringent test of whether video-based adaptation can provide effective 3D generalization.

With the proposed hybrid supervision, \name~consistently improves both MV-DUSt3R and VGGT across all four benchmarks. 
For MV-DUSt3R, \name~(10K) reduces CD by 6\%, 20\%, 29\%, and 42\% on ScanNet, 7Scenes, TUM-RGBD, and MP3D, respectively, and further brings a substantial DAc gain of +17.7 on MP3D. 
For VGGT, \name~also improves both DAc and CD on every dataset, showing that our adaptation strategy is effective across different pretrained 3D backbones.

Notably, on the zero-shot benchmarks 7Scenes and TUM-RGBD, MV-DUSt3R+\name~matches or surpasses the strongest fully 3D-supervised foundation models on both DAc and CD, achieving 98.0/0.76 on 7Scenes and 93.1/1.85 on TUM-RGBD. 
On ScanNet and MP3D, models trained with substantially larger curated 3D corpora still retain advantages in some metrics. 
This suggests that video-based adaptation is not a replacement for large-scale 3D pretraining, but rather an orthogonal and potentially composable direction for improving pose-free 3D foundation models.

We also compare with VDS, a representative video-based baseline that fine-tunes MV-DUSt3R using pseudo depth from Depth Any Video~\cite{yang2024depthanyvideo}. While VDS slightly improves in-distribution performance, it degrades zero-shot generalization on TUM-RGBD and MP3D, indicating that noisy and temporally inconsistent pseudo-depth can harm geometric representations. This supports the need for \name's SfM-anchored and differentiable consistency design. Qualitative comparisons are provided in App.~\ref{app_vis}.

\begin{table}[t]
\centering
\footnotesize
\caption{\textbf{Ablation study on hybrid supervision and training settings.} We evaluate the contribution of each supervision component and the impact of different training strategies.}
\label{tab:ablation}
\resizebox{0.85\columnwidth}{!}{
\begin{tabular}{lcccccccc}
\toprule
& \multicolumn{2}{c}{ScanNet} & \multicolumn{2}{c}{7Scenes} & \multicolumn{2}{c}{TUM} & \multicolumn{2}{c}{MP3D} \\
\cmidrule(lr){2-3} \cmidrule(lr){4-5} \cmidrule(lr){6-7} \cmidrule(lr){8-9}
Method & DAc$\uparrow$ & CD$\downarrow$ & DAc$\uparrow$ & CD$\downarrow$ & DAc$\uparrow$ & CD$\downarrow$ & DAc$\uparrow$ & CD$\downarrow$ \\
\midrule
Baseline (Foundation) & 88.2 & 1.79 & 89.6 & 0.95 & 87.5 & 2.60 & 30.3 & 43.2 \\
\rowcolor[HTML]{EFEFEF} 
\textbf{Ours (Full)} & \textbf{91.2} & \textbf{1.69} & \textbf{98.0} & \textbf{0.76} & \textbf{93.1} & \textbf{1.85} & \textbf{48.0} & \textbf{24.9} \\
\midrule
\textit{Component Ablation:} \\
w/o Sparse Anchors ($\mathcal{L}_{\mathrm{anchor}}$) & 77.0 & 4.33 & 88.5 & 1.43 & 69.4 & 5.20 & 18.9 & 33.6 \\
w/o Dense Consistency ($\mathcal{L}_{\text{ren}}$) & 89.2 & 1.89 & 93.5 & 0.79 & 86.1 & 2.26 & 49.2 & 26.0 \\
w/o Supervised Replay & 76.0 & 3.32 & 88.5 & 1.25 & 80.6 & 2.14 & 45.4 & 32.5 \\
w/o Frozen GS Params & 91.2 & 1.70 & 93.5 & 0.72 & 91.7 & 2.09 & 46.7 & 23.7 \\
\midrule
\textit{Strategy Variants:} \\
w/ Linear Sampling (vs. Jitter) & 91.0 & 1.88 & 95.0 & 0.84 & 90.2 & 2.06 & 49.0 & 25.1 \\
w/ 8 Target Views (vs. 2) & 90.2 & 1.73 & 96.5 & 0.78 & 90.3 & 1.84 & 49.7 & 26.0 \\
\bottomrule
\end{tabular}
}
\end{table}
\vspace{-5pt}

\subsection{Ablations}
\paragraph{Effect of Hybrid Supervision}
We evaluate the contribution of each component in our hybrid training strategy by removing one supervision at a time. Specifically, we independently ablate the Sparse Anchors, the Dense Consistency, and the small amount of supervised replay data used to stabilize fine-tuning. The results in Table~\ref{tab:ablation} show that the full combination achieves the best reconstruction quality, while excluding any single component leads to a noticeable performance drop.
We further examine whether to freeze the pretrained Gaussian representation module during fine-tuning. Results indicate that freezing the GS module leads to more stable optimization and better adaptation of the reconstruction network, whereas updating GS jointly with the network tends to degrade performance.

\paragraph{Number of Rendered Views}
We analyze the effect of varying the number of rendered novel views per training sample, as shown in Table~\ref{tab:ablation}. By default, we supervise the model with two target views to balance training efficiency and supervision quality. Increasing the number of rendered views (e.g., w/ 8 Target Views) provides stronger geometric constraints but also exposes the model to more unseen regions across views, which may introduce noise into the supervision signal. 
% Empirically, using 8 target views not only yields better reconstruction results on indoor datasets, but also leads to noticeably improved generalization on outdoor scenes.
\paragraph{Frame Sampling Strategy}
We ablate the effect of our perturbed frame sampling strategy against linear sampling. As shown in Table~\ref{tab:ablation}, compared with video frame sampling without perturbation (w/ Linear Sampling), our perturbed sampling consistently yields higher accuracy and more stable performance across benchmarks. These improvements validate its effectiveness, and we adopt it as the default strategy for constructing multi-view training sequences.
We further compare against content-aware context and target selection based on rotation-compensated flow and view overlap (App.~\ref{app:content_sampling}). Temporal jitter remains a strong accuracy--simplicity trade-off on RE10K/DL3DV, whereas content-aware selection is most beneficial on wide-baseline MP3D (CD 24.9$\to$23.4), suggesting its value for less regular camera motion.

\section{Discussion}

\subsection{Beyond Scale: How Does the Data Difficulty Spectrum Shape Generalization?}
% {How to Select Data for Adaptation?}
\begin{figure}[t]
\centering
\includegraphics[width=0.7\textwidth]{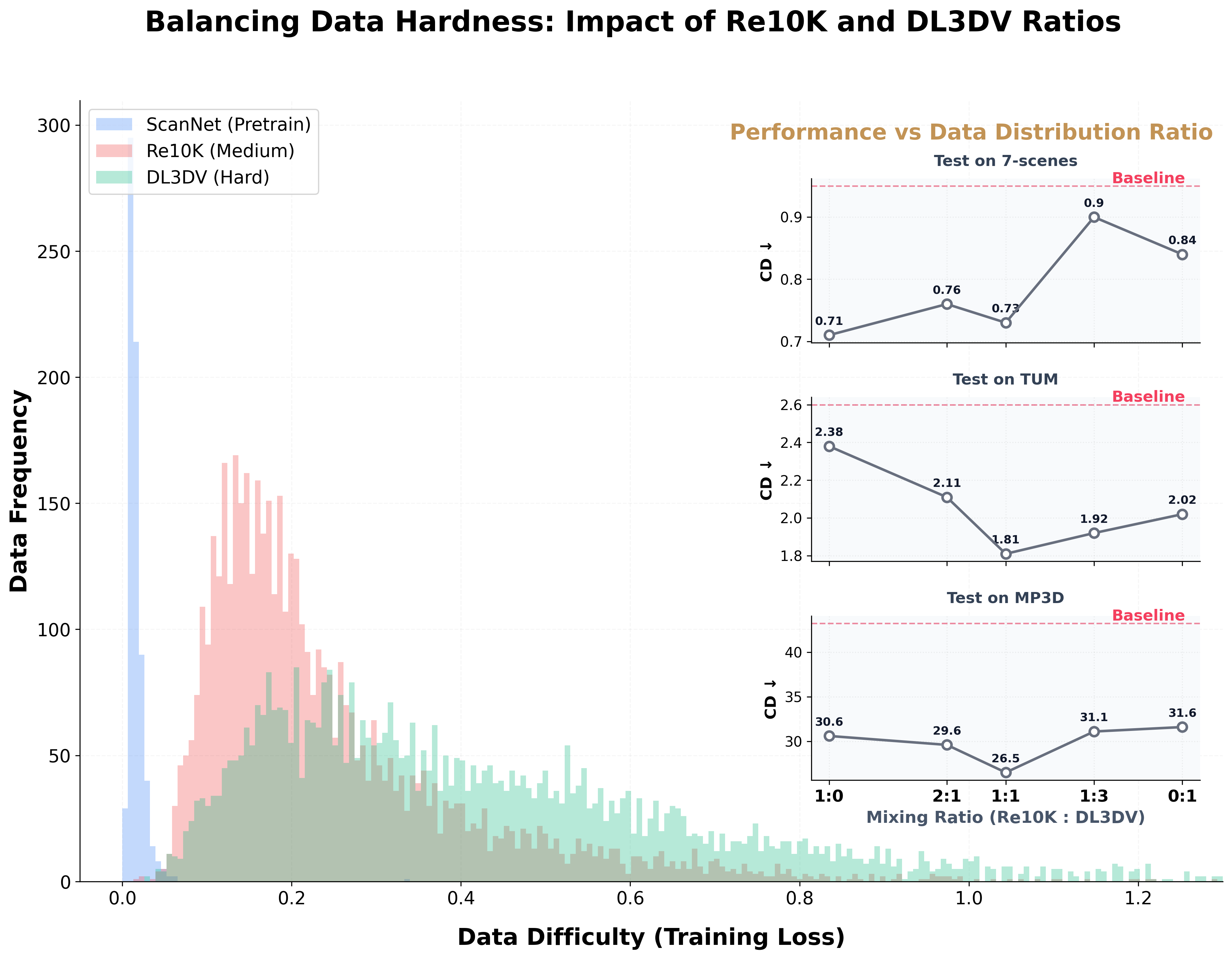}
\caption{\footnotesize{Empirical analysis of training data difficulty and its impact on model generalization.
(Left) Training distribution across three datasets. (Right) Generalization performance (CD $\downarrow$) on three test sets under various mixing ratios of Re10K to DL3DV samples. Subplots indicate that while pure medium-hardness data (1:0) benefits simple scenes, a balanced mixture (1:1) yields the most robust generalization on large-scale environments like MP3D. The dashed red lines denote the pre-trained baseline performance.}}
\label{fig:data_distribution}
\end{figure}
We analyze the training data distribution in terms of sample difficulty, measured by the per-sample training loss of the pretrained model. As shown in the left panel of Figure~\ref{fig:data_distribution}, the Re10K dataset predominantly contains medium-difficulty samples, while DL3DV offers a higher proportion of challenging samples with elevated training losses. This observation aligns with our intuition, as Re10K features more structured indoor scenes, whereas DL3DV encompasses complex real-world environments with diverse lighting conditions and cluttered backgrounds.
As shown in the right panel of Figure~\ref{fig:data_distribution}, we conduct a sensitivity analysis by varying the mixing ratio between medium-difficulty samples (Re10K) and high-difficulty samples (DL3DV) to investigate how the difficulty composition of training data influences cross-dataset generalization. The pre-trained baseline (red dashed line) consistently exhibits the highest error across all benchmarks, underscoring the necessity of fine-tuning on diverse data. For relatively structured and simple scenes, such as 7Scenes, the model achieves its peak performance with a higher proportion of medium-difficulty data (\eg, the 1:0 ratio). In contrast, for large-scale and complex environments like MP3D and TUM, the inclusion of high-difficulty samples is crucial for pushing the performance boundary. As shown in the right panels of Figure~\ref{fig:data_distribution}, the Chamfer Distance (CD) significantly drops as the DL3DV ratio increases, reaching its optimum at the balanced 1:1 ratio. Beyond this point, however, we observe diminishing returns or slight performance fluctuations. These results suggest that a balanced curriculum of medium and high difficulty is essential for robust generalization.

\subsection{Regularization Strength \& Prevention of Forgetting}
\begin{table}[t]
\centering
\small
\setlength{\tabcolsep}{5pt}
\caption{\textbf{Effect of Video-to-Supervised Data Ratio.}
Evaluation on four datasets with different video-to-supervised sample ratios. A 30:1 ratio is adopted as the default, balancing reconstruction quality with reliance on pretrained supervised data.}
\label{tab:supervision_ratio}
\begin{tabular}{cccccccccc}
  \toprule
  & \multicolumn{2}{c}{ScanNet} & \multicolumn{2}{c}{7Scenes} & \multicolumn{2}{c}{TUM} & \multicolumn{2}{c}{MP3D} \\
  \cmidrule(lr){2-3} \cmidrule(lr){4-5} \cmidrule(lr){6-7} \cmidrule(lr){8-9}
  Ratio & DAc$\uparrow$ & CD$\downarrow$ & DAc$\uparrow$ & CD$\downarrow$ & DAc$\uparrow$ & CD$\downarrow$ & DAc$\uparrow$ & CD$\downarrow$ \\
  \midrule
  10:1  & 90.0 & 1.77 & 94.5 & 0.75 & 91.6 & 1.84 & 44.9 & 24.55 \\
  20:1  & 91.2 & 1.62 & 98.0 & 0.76 & 89.0 & 2.08 & 47.2 & 24.90 \\
  \rowcolor[rgb]{.906,.902,.902}
  30:1  & 91.2 & 1.69 & 98.0 & 0.76 & 93.1 & 1.85 & 48.0 & 24.90 \\
  100:1 & 90.0 & 1.84 & 94.5 & 0.87 & 93.0 & 1.89 & 43.4 & 23.73 \\
  \bottomrule
\end{tabular}
\end{table}
%As discussed in our method, incorporating a small amount of strongly supervised data from the pretraining stage is crucial for stabilizing training. To analyze its impact, we vary the ratio between video-based data and pretrained supervision data to construct different supervision settings. As shown in Table~\ref{tab:supervision_ratio}, a 30:1 ratio achieves the best overall performance across datasets, while even a 100:1 ratio still yields competitive results. This indicates that only a small amount of supervised data is sufficient to serve as effective anchors during training.
As formulated in Eq.~\ref{eq:loss_all}, the supervised replay data acts as a regularization term. A key question is: \textit{what is the minimal regularization strength required to prevent catastrophic forgetting?}
To answer this, we vary the ratio between weakly-supervised video data $\mathcal{D}_{video}$ and supervised replay data $\mathcal{D}_{sup}$.
As shown in Table~\ref{tab:supervision_ratio}, a $30:1$ ratio achieves the optimal balance. Without sufficient regularization (\eg, ratio $\to \infty$), the model drifts from the metric scale derived from ScanNet. Conversely, even a very sparse signal (100:1) significantly stabilizes the optimization, confirming that the foundation model requires only lightweight anchoring to maintain its priors while scaling up.

\subsection{Capability Gains from Video-based Adaptation}
\paragraph{Pose Estimation}
We observe that the proposed training scheme with video data not only improves the quality of the reconstruction, but also improves the accuracy of the model pose estimation. Following MV-DUSt3R ~\cite{tang2024mvdust3r}, we estimate camera poses using a standard RANSAC~\cite{ransac} and PnP~\cite{pnp} pipeline, where 2D--3D correspondences are formed between image pixels and the predicted pointmaps generated by the network. In Table~\ref{tab:pose_estimation}, our \name~yields consistently lower rotation and translation errors across multiple datasets, demonstrating that video-based fine-tuning leads to more geometrically reliable outputs even without access to ground-truth annotations of video.
\paragraph{Evaluation on OOD Outdoor Scenes}
Although the training videos primarily consist of indoor scenes, they also include a portion of outdoor scenes. As shown in Table~\ref{tab:comparison_results}, \name~generalizes surprisingly well to the out-of-distribution outdoor UASOL dataset~\cite{bauer2019uasol}, achieving a DAc@0.5 of 27.8 and outperforming the baseline MV-DUSt3R~\cite{tang2024mvdust3r}. This demonstrates the advantage of leveraging diverse video data, where incidental exposure to outdoor environments contributes to improved robustness beyond the training distribution.

\begin{table}[t]
\centering
\captionsetup{font=footnotesize}
\begin{minipage}[b]{0.53\textwidth}
\centering
\footnotesize
\setlength{\tabcolsep}{1pt}
\renewcommand{\arraystretch}{0.95}
\caption{\textbf{Impact of Fine-tuning with Videos on Pose Estimation.}
Comparison between the baseline and our method on relative rotation error (RRE), relative translation error (RTE), and mean average error (mAE). Lower is better.}
\label{tab:pose_estimation}
\begin{tabular}{c|ccc|ccc|ccc}
  \toprule
  & \multicolumn{3}{c|}{7Scenes} & \multicolumn{3}{c|}{TUM RGBD} & \multicolumn{3}{c}{MP3D} \\
  Method & RRE & RTE & mAE & RRE & RTE & mAE & RRE & RTE & mAE \\
  \midrule
  MV-DUSt3R    & 0.3 & 3.2 & 13.8 & 2.7 & 16.0 & 27.0 & 56.9 & 55.9 & 67.3 \\
  \rowcolor[rgb]{.906,.902,.902}
  Ours (\name) & 0.2 & 2.9 & 13.3 & 2.0 & 15.7 & 26.4 & 41.2 & 42.5 & 55.9 \\
  \bottomrule
\end{tabular}
\end{minipage}
\hfill
\begin{minipage}[b]{0.46\textwidth}
\centering
\footnotesize
\setlength{\tabcolsep}{2pt}
\renewcommand{\arraystretch}{0.95}
\caption{\textbf{Reconstruction on UASOL (out-of-distribution, outdoor dataset).}
\name~outperforms MV-DUSt3R and the VDS variant, which suffers from inconsistent and noisy multi-frame depth predictions.
% Despite utilizing more video data, the pseudo-supervised model suffers from inconsistent and noisy multi-frame depth predictions.
}
\label{tab:comparison_results}
\begin{tabular}{lcc}
  \toprule
  \textbf{Method} & \textbf{DAc-0.5} $\uparrow$ & \textbf{CD} $\downarrow$ \\
  \midrule
  MV-DUSt3R & 3.3 & 27.8 \\
  MV-DUSt3R + VDS & 26.0 & 31.8 \\
  Ours (\name) & \textbf{27.8} & \textbf{18.8} \\
  \bottomrule
\end{tabular}
\end{minipage}
\end{table}

% \paragraph{NVS}

% \paragraph{Multi-view Recon}

% \input{sec/5_limitation}
\section{Conclusion}
\label{sec:Conclusion}
% We present \name, a framework that leverages large-scale, diverse Internet video sequences to fine-tune pretrained 3D reconstruction models. By combining a perturbed frame sampling strategy with hybrid supervision (3D sparse points and 2D novel view consistency), \name~effectively adapts models to diverse scenes. Extensive experiments across multiple benchmarks demonstrate consistent and substantial improvements over strong baselines, highlighting the potential of raw videos as a scalable and effective source of supervision for boosting 3D reconstruction model.

We introduce \textbf{SAGE}, a framework that unlocks the scalability of geometric foundation models by exploiting the intrinsic consistency of Internet video. 
By formulating adaptation as a regularized weakly-supervised optimization, SAGE balances structural bias (from sparse anchors) and variance reduction (from dense differentiable consistency) to robustly learn from noisy data.
Our experiments show log-linear scaling laws in geometric performance, reducing zero-shot Chamfer Distance by \mbox{20--42\%} on MV-DUSt3R with consistent gains on VGGT. 
This work establishes a scalable paradigm for evolving general-purpose 3D models using abundant static-scene video, turning the data scarcity bottleneck into a data abundance opportunity.

\begin{ack}
This work was in part supported by  National Natural Science Foundation of China (No.~62576315), National Research Foundation, Singapore under its National Research Foundation Fellowship in Artificial Intelligence (NRF Award NRFFIAI1-2024-0010),  NTU FA-SUG,   Ningbo Global Innovation Center, Zhejiang University (Grant No.~701G26003), and  Hangzhou National AI Application Pilot Base (Embodied Intelligence).
%
%\textbf{Competing interests.} 
The authors declare no competing interests.
\end{ack}

\clearpage
\bibliography{example_paper}
\bibliographystyle{icml2026}

\appendix
\clearpage
\onecolumn
\appendix

\etocdepthtag.toc{mtappendix}
\etocsettagdepth{mtchapter}{none}
\etocsettagdepth{mtappendix}{subsection}

%	\part{Appendix} % Start the appendix part
%	\renewcommand{\contentsname}{}
\renewcommand{\contentsname}{Appendix}
% \tableofcontents

% \usetikzlibrary{shapes.geometric, arrows.meta, positioning, calc, shadows}

% ICML-style color palette
\definecolor{processblue}{RGB}{232, 241, 250}
\definecolor{darkblue}{RGB}{50, 100, 150}
\definecolor{lightgray}{RGB}{245, 245, 245}

% \clearpage
%%%%%%%%% BODY TEXT - ENTER YOUR RESPONSE BELOW

\section{Preliminary}

% \subsection{Common Paradigm of 3D Geometric Foundation Models}
% Recent 3D geometric foundation models (3D GFMs), represented by DUSt3R~\cite{wang2024dust3r} and VGGT~\cite{wang2025vggt}, follow a largely unified pipeline that performs end-to-end geometric reasoning from RGB images, shown in Figure~\ref{fig:GFM_pipeline}. These models first encode one or multiple input views into visual tokens using a shared image encoder, followed by cross-view feature interaction—typically implemented via correspondence-aware attention or transformer-based aggregation—to capture multi-view geometric consistency. Geometry is then directly regressed in the form of pixel-aligned 3D point maps, depth, and/or camera parameters. DUSt3R formulates reconstruction as a pairwise image matching problem, which offers robustness under minimal view settings but limits scalability to long sequences, whereas VGGT extends this paradigm to explicit multi-view modeling with higher training resolution and stronger geometric supervision, achieving improved performance at increased computational cost. Despite architectural differences, most existing 3D GFMs rely heavily on large-scale, strongly supervised 3D datasets for pretraining, which constrains their scalability and motivates alternative adaptation strategies using weaker but more scalable supervision sources.
\subsection{Common Paradigm of 3D Geometric Foundation Models}
\begin{figure}[ht]
\centering
\includegraphics[width=0.6\textwidth]{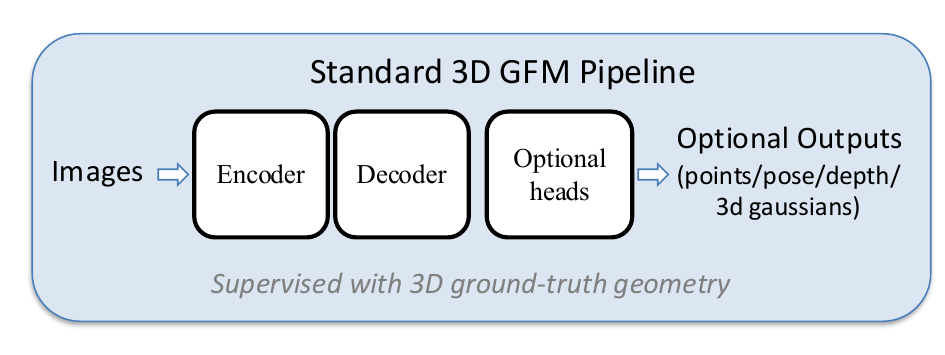}
\caption{\textbf{Standard 3D GFM Training and Inference Framework.}
A generic 3D GFM consists of an image encoder, a decoder for cross-view feature interaction, and optional output heads that regress various geometric representations. These models are commonly trained with explicit 3D supervision.
}
\label{fig:GFM_pipeline}
\end{figure}

Recent 3D geometric foundation models (3D GFMs), represented by DUSt3R~\cite{wang2024dust3r} and VGGT~\cite{wang2025vggt}, follow a largely unified pipeline that performs end-to-end geometric reasoning from RGB images, as illustrated in Figure~\ref{fig:GFM_pipeline}. 
Given one or multiple input views, these models first extract visual tokens with a shared image encoder, then perform cross-view feature interaction to capture multi-view geometric consistency, and finally regress geometry in pixel-aligned forms such as 3D point maps, depth maps, and camera parameters.

We instantiate this paradigm with two representative backbones that differ in their cross-view interaction and output parameterization. 
MV-DUSt3R extends DUSt3R to multiple views using an encoder--decoder architecture with cross-view cross-attention, and predicts point maps in a selected reference frame. 
VGGT, in contrast, uses a single transformer with alternating frame-wise and global self-attention, together with multi-task heads for point maps, depth, and camera parameters. 
These complementary designs allow us to evaluate whether our adaptation strategy transfers across both reference-frame point-map reconstruction and unified multi-view transformer prediction. 
Despite such architectural progress, existing 3D GFMs still rely heavily on large-scale, strongly supervised 3D datasets for pretraining, which constrains their scalability and motivates alternative adaptation strategies using weaker but more scalable supervision sources.

\section{Experimental Details}
\subsection{Video Data Processing Pipeline}
\label{app:data_pocess}
\begin{figure}[ht]
\centering
\includegraphics[width=0.75\textwidth]{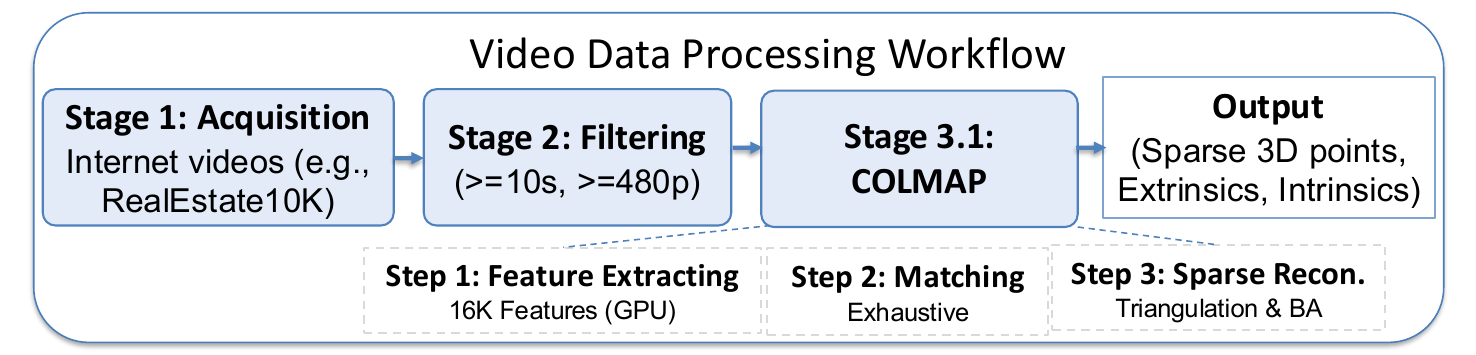}
\caption{Video Data Processing Workflow for Weak Geometric Supervision.
Raw Internet videos are filtered based on duration and resolution and processed using a COLMAP-based structure-from-motion pipeline. Feature extraction, exhaustive matching, and sparse reconstruction are performed to obtain sparse 3D points and camera parameters, which serve as weak geometric supervision for scalable model adaptation.}
\label{fig:data_flow}
\end{figure}

This section details the video data processing pipeline used to construct the training set for weakly-supervised adaptation of 3D geometric foundation models. An overview of the workflow is illustrated in Fig.~\ref{fig:data_flow}.

\textbf{1) Video Acquisition.}
We collect raw videos from large-scale Internet video datasets, RealEstate10K and DL3DV contain minimal dynamic content, making them well-suited for fine-tuning pretrained 3D reconstruction models. These videos provide rich multi-view observations while avoiding reliance on curated 3D ground-truth geometry.
\textbf{2) Video Filtering.}
To ensure sufficient geometric content and temporal coverage for reliable multi-view reconstruction, we further filter out low-quality video sequences, those are too short, exhibit large viewpoint discontinuities, or contain dynamic objects, as they may cause COLMAP reconstruction failures or produce excessively noisy point clouds: (1) minimum duration of 10 seconds, and (2) minimum spatial resolution of 480p. Videos that do not meet these requirements are discarded. This filtering step removes low-quality or short clips that are unlikely to yield stable camera poses or consistent multi-view correspondences.
\textbf{3) Structure-from-Motion via COLMAP.}
For each retained video, we run a standard Structure-from-Motion (SfM) pipeline using COLMAP to recover sparse 3D geometry and camera parameters. Specifically, we first extract up to 16K local features per frame using GPU acceleration, which improves robustness under challenging conditions such as motion blur, illumination changes, and textureless regions. We then perform exhaustive feature matching across frames to maximize reliable correspondences and support wide-baseline reconstruction. Finally, COLMAP performs triangulation and bundle adjustment to estimate sparse 3D point clouds, camera extrinsic parameters, and camera intrinsic parameters. Only reconstructions that successfully converge are retained for subsequent training.
\textbf{4) Output Representation.}
The final output of the pipeline consists of sparse 3D points with associated camera intrinsics and extrinsics, which serve as weak geometric supervision for model adaptation. Notably, no dense depth maps or ground-truth meshes are used at any stage of training. This design allows the proposed method to scale to large and diverse Internet video collections while maintaining compatibility with standard SfM tools.

\paragraph{Pipeline statistics at scale.}
To quantify the practical scalability of our preprocessing pipeline, we report the following statistics measured on the RE10K subset used for training.
The high success rate benefits from RE10K's upstream pre-filtering: clips with motion blur or aggressive viewpoint discontinuity are largely already excluded by the dataset curators. Crucially, COLMAP failures produce \textit{missing} data rather than \textit{erroneous} data: clips that do not converge are simply discarded, so no noisy supervision is injected into training. The end-to-end preprocessing for our $10\text{K}{+}$ scene corpus is a modest one-time cost that runs in parallel on commodity GPUs.
\begin{table}[h]
\centering
\small
\setlength{\tabcolsep}{6pt}
\caption{\textbf{COLMAP preprocessing statistics on the RealEstate10K subset.}
The pipeline scales to thousands of clips with negligible failure rate; preprocessing for the entire $10\text{K}{+}$ scene set is a one-time, modest cost.}
\label{tab:colmap_stats}
\begin{tabular}{ll}
\toprule
\textbf{Statistic} & \textbf{Value} \\
\midrule
RE10K clips processed     & $>$13K (50--200 frames per clip) \\
COLMAP failures           & 8 clips ($<\!0.1\%$) \\
Average processing time   & ${\sim}15$ sec / clip (170 frames), GPU-accelerated \\
Sparse points / frame     & 300--2000 (mean ${\sim}1500$) \\
Total preprocessing time  & ${\sim}$5--6 h for $>$13K clips (4$\times$ RTX 4090, 2 workers) \\
Reconstruction stage      & sparse SfM only (no dense MVS) \\
\bottomrule
\end{tabular}
\end{table}

\subsection{Experimental Setup}
\label{app_details}
\paragraph{Datasets}
We use videos from RealEstate10K~\cite{zhou2018re10k} and DL3DV~\cite{ling2024dl3dv} as the primary training sources. These datasets contain a large number of diverse scenes of common real-life environments. 
%We first preprocess all video clips, 
%and during training, samples are randomly selected from the preprocessed set at each iteration. 
We preprocess the dataset by filtering out short videos. 
In total, over 10,000 video clips are utilized for training, resulting in a 10-fold increase in scene diversity compared to standard 3D datasets.
To assess the reconstruction performance, we evaluate all models on four benchmarks. ScanNet~\cite{dai2017scannet} is divided into training and test sets; all methods are trained on the training split and tested on the official test split. In addition, we include 7Scenes~\cite{shotton20137scenes}, TUM RGB-D~\cite{tum}, and Matterport3D~\cite{chang2017matterport3d} as zero-shot datasets, which are not seen during training by any of the methods. These datasets provide a more rigorous assessment of generalization ability to unseen and diverse real-world scenes.

\paragraph{Evaluation metrics}
We evaluate reconstruction quality using two main metrics: Chamfer Distance (CD)~\cite{chamfer_distance} and Distance Accuracy @0.2 (DAc), which measures the proportion of pixels whose normalized distance to the ground-truth pointmap is less than or equal to 0.2. CD serves as a global geometric error metric but is sensitive to point cloud density and outliers, and can be affected by prediction sparsity or noise. In contrast, DAc directly quantifies the proportion of well-aligned points and more reliably reflects overall structural consistency, making it particularly suitable for evaluating dense and accurate reconstructions.
In more challenging settings, such as the MP3D~\cite{chang2017matterport3d} dataset characterized by low viewpoint overlap and textureless surfaces, we use Distance Accuracy @0.5 (DAc@0.5) as a more appropriate metric. It offers a more reliable assessment of reconstruction performance under such conditions.

\subsection{Implementation details}

\paragraph{Video Frame Sampling}
When constructing training inputs, we follow the uniform sampling strategy described in the main text. For RealEstate10K~\cite{zhou2018re10k}, we set the temporal interval $\Delta t = 20$ based on its typical recording frame rate, while for DL3DV~\cite{ling2024dl3dv}, we use $\Delta t = 15$ due to its faster motion and denser frame sampling. For the perturbed sampling variant, we apply a random offset of up to $\pm 5$ frames around each sampled timestamp to increase viewpoint diversity.
\paragraph{Test Dataset Construction}
Our test dataset construction follows the protocol proposed in CroCo~\cite{weinzaepfel2022croco} and MV-DUSt3R~\cite{tang2024mvdust3r}, but with key modifications.
Unlike MV-DUSt3R, we do not use synthetic images rendered via Habitat-Sim~\cite{szot2021habitat}, as such rendered data often suffers from limited photorealism and domain discrepancies compared to real-world images.
Instead, for all test datasets, we begin by randomly selecting the first image in each test sequence. Subsequent views are sampled based on their geometric overlap with the previously selected views, which is computed using ground-truth depth. A predefined threshold is applied to determine whether a candidate frame provides sufficient overlap to be included in the test sequence.
\paragraph{Training Process}
\label{app:train_detail}
When we finetune the MV-DUSt3R, the resolution of input images is $224 \times 224$. Each training sample consists of 8 input views, with either 2 or 8 novel views rendered for supervision. During training, we mix the video data from RealEstate10K and DL3DV. Additionally, we include approximately 3\% (about 400) pretrained samples, resulting in a 30:1 ratio between video data and pretrained supervision data.
To preserve the pretrained 3D Gaussian representation, we freeze the gaussian head and only optimize the reconstruction network. The model is trained for 5 epochs. In each epoch, we sample 15,000 training trajectories from each dataset. The entire training process is highly efficient, requiring less than two days on 4 RTX 4090 GPUs using only low-cost video data.
As for the VGGT finetuning, we follow the original VGGT resolution setting and the model is trained for 15 epochs. Since VGGT has no Gaussian head, we attach a lightweight Gaussian head that regresses opacity, scale, rotation, and color, with the Gaussian centers given by VGGT's predicted 3D points; the head is trained for 2--3 epochs and then frozen during adaptation.

\section{Additional Experimental Results}
% \subsection{Generalization to Other 3D Foundation Model}
% To further verify the generality of our training framework, we apply \name~to another multi-view feedforward 3D reconstruction model, VGGT~\cite{wang2025vggt}.
% Specifically, we fine-tune VGGT on a subset of RealEstate10K for approximately five hours on 4 GPUs, following the original training resolution of 518.
% As shown in Table~\ref{tab:vggt}, fine-tuning with \name~reduces Chamfer Distance across all datasets, with particularly large improvements of up to 26\% on 7Scenes and TUM-RGBD.
% \begin{table}[htbp]
%   \centering
%   % \small
%   \setlength{\tabcolsep}{2pt}
%   \caption{\textbf{Performance of VGGT before and after fine-tuning with \name~on four 3D reconstruction benchmarks.} Chamfer Distance (CD) decreases consistently, with particularly large improvements on 7Scenes and TUM-RGBD.}
% \begin{tabular}{cccccccccc}
%     \toprule
%     % \toprule
%     & \multicolumn{2}{c}{ScanNet} & \multicolumn{2}{c}{7Scenes} & \multicolumn{2}{c}{TUM} & \multicolumn{2}{c}{MP3D} \\
%     \cmidrule(lr){2-3} \cmidrule(lr){4-5} \cmidrule(lr){6-7} \cmidrule(lr){8-9}
%     Method & DAc↑ & CD↓ & DAc↑ & CD↓ & DAc↑ & CD↓ & DAc↑ & CD↓ \\
%     \midrule
%     VGGT  & \textbf{90.0} & 3.12 & 99.0 & 0.65 & 80.0 & 3.92 & \textbf{57.0} & 20.36 \\    
%     VGGT+\name  & 88.0  & \textbf{2.79}  & \textbf{100.0}  & \textbf{0.48}  & \textbf{82.9}  & \textbf{2.89}  & 55.9  & \textbf{18.21} \\
%     \bottomrule
%     % \bottomrule 
%   \end{tabular}
%     \vspace{-3mm}
%   \label{tab:vggt}
% \end{table}

\subsection{Multi-view Reconstruction on ScanNet}
\label{app:multi-view}
In the main paper, we report 3D point cloud reconstruction results using 8 input views. Here, we provide additional results under varying numbers of input views. Table~\ref{tab:scannet_scene_results} shows the performance on the ScanNet test set across different view counts. As shown in Table~\ref{tab:scannet_scene_results}, our method achieves the best overall reconstruction performance across most settings. However, as the number of input views increases, single forward inference becomes more challenging. 
% In such cases, optimization-based methods like DUSt3R~\cite{wang2024dust3r} can benefit from global alignment in the backend to achieve improved results. 
We also report more results on MP3D~\cite{chang2017matterport3d} with varying number of inputs in Table~\ref{tab:mp3d_scene_results}. For multi-view reconstruction,
Spann3R~\cite{wang2024spann3r} and DUSt3R~\cite{wang2024dust3r} are excluded due to high memory usage and optimization overhead, respectively. Only single forward inference methods are compared.
\vspace{-8pt}
\begin{table}[htbp]
  \centering
  % \footnotesize
  \caption{\textbf{Multi-view 3D reconstruction performance on ScanNet.} 
  Comparison across five ScanNet scenes with varying number of images (4 to 24).
  Each column reports Distance Accuracy (DAc$\uparrow$) and Chamfer Distance (CD$\downarrow$).}
  \setlength{\tabcolsep}{2pt}
  \begin{tabular}{ccccccccccc}
    \toprule
    \toprule
    \multirow{2}{*}{Method} &  \multicolumn{2}{c}{4} & \multicolumn{2}{c}{12} & \multicolumn{2}{c}{16} & \multicolumn{2}{c}{20} & \multicolumn{2}{c}{24} \\
    \cmidrule(lr){2-3} \cmidrule(lr){4-5} \cmidrule(lr){6-7} \cmidrule(lr){8-9} \cmidrule(lr){10-11}
    & DAc & CD & DAc & CD & DAc & CD & DAc & CD & DAc & CD \\
    \midrule
    % DUSt3R     & 87.3 & 3.92 & 82.4 & 2.95 & 80.4 & 2.91 & 79.4 & 2.71 & \textbf{80.4} & \textbf{2.26} \\
    VGGT        & 72.0 & 5.19 & 58.0 & 4.83 & 50.0 & 4.91 & 45.0 & 5.56 & 36.0 & 5.62 \\
    MV-DUSt3R & 88.0 & \textbf{1.67} & \textbf{86.0} & 2.63 & 82.0 & 2.43 & 80.0 & 2.95 & \textbf{73.0} & 2.49 \\
    Ours                    & \textbf{90.0} & 1.71 & 84.0 & \textbf{2.31} & \textbf{82.0} & \textbf{1.98} & \textbf{82.0} & \textbf{2.41} & 72.0 & \textbf{2.34} \\
    \bottomrule
    \bottomrule
  \end{tabular}
  \label{tab:scannet_scene_results}
\end{table}

% \subsection{Multi-view Reconstruction on MP3D}

\begin{table}[htbp]
  \centering
  % \footnotesize
  \caption{\textbf{Multi-view 3D performance on MP3D.} Evaluation across different number of images.
  Each column reports Distance Accuracy (DAc@0.5$\uparrow$) and Chamfer Distance (CD$\downarrow$).
  All methods are evaluated without global alignment.}
  \setlength{\tabcolsep}{1.5pt}
  \begin{tabular}{ccccccccccc}
    \toprule
    \toprule
    \multirow{2}{*}{Method} & \multicolumn{2}{c}{4} & \multicolumn{2}{c}{12} & \multicolumn{2}{c}{16} & \multicolumn{2}{c}{20} & \multicolumn{2}{c}{24} \\
    \cmidrule(lr){2-3} \cmidrule(lr){4-5} \cmidrule(lr){6-7} \cmidrule(lr){8-9} \cmidrule(lr){10-11}
    & DAc & CD & DAc & CD & DAc & CD & DAc & CD & DAc & CD \\
    \midrule
    VGGT      & 44.1 & 40.65 & 24.6 & 46.28 & 19.5 & 57.45 & 13.3 & 57.56 & 13.3 & 78.27 \\
    MV-DUSt3R & 42.9 & 42.08 & 23.0 & 49.28 & 24.0 & 62.29 & 11.7 & 77.89 & 7.1 & 92.19 \\
    \name~(Ours)                    & \textbf{62.2} & \textbf{19.28} & \textbf{35.2} & \textbf{30.41} & \textbf{30.1} & \textbf{34.73} & \textbf{18.9} & \textbf{39.69} & \textbf{14.3} & \textbf{52.61} \\
    \bottomrule
    \bottomrule
  \end{tabular}
  \label{tab:mp3d_scene_results}
\end{table}

% \subsection{More Results on BlendedMVS}

\subsection{Effect of Viewpoint Interpolation for Novel View Supervision}
As discussed in the main paper, when constructing novel view supervision, we observe that interpolated target views (views that lie spatially between the input images) consistently yield better results than extrapolated ones. This is because extrapolated views tend to fall outside the effective coverage of the input images, making view consistency supervision more susceptible to ambiguity and error, which is also analyzed in Splatt3r~\cite{smart2024splatt3r}. In contrast, interpolated views benefit from overlapping content on both sides, leading to more stable and effective training signals. In our experiments, we provide a comparative analysis of both strategies. 

We ablate the effect of target view selection for novel-view supervision. As shown in Table~\ref{tab:nvs_interpolation}, interpolated views consistently yield better reconstruction performance compared to extrapolated ones under the same training setup.
Note that this experiment is conducted under a slightly different configuration from the main paper and serves primarily for comparative analysis.
\begin{table}[htbp]
  \centering
  % \small
  \caption{\textbf{Comparison of novel-view supervision strategies.}
  Performance across four datasets using extrapolated vs.\ interpolated target views.
  Interpolated targets consistently yield better reconstruction results.}
  \setlength{\tabcolsep}{3pt}
  \begin{tabular}{ccccccccc}
    \toprule
    \toprule
    \multirow{2}{*}{Method} & \multicolumn{2}{c}{ScanNet} & \multicolumn{2}{c}{7Scenes} & \multicolumn{2}{c}{TUM} & \multicolumn{2}{c}{MP3D} \\
    \cmidrule(lr){2-3} \cmidrule(lr){4-5} \cmidrule(lr){6-7} \cmidrule(lr){8-9}
    & DAc & CD & DAc & CD & DAc & CD & DAc & CD \\
    \midrule
    Ours (extra) & 88.0 & 2.31 & 100.0 & 0.74 & 87.5 & 2.53 & 30.6 & 39.91 \\
    Ours (inter) & \textbf{90.0} & \textbf{1.88} & \textbf{100.0} & \textbf{0.48} & \textbf{91.7} & \textbf{1.78} & \textbf{46.4} & \textbf{21.80} \\
    \bottomrule
    \bottomrule
  \end{tabular}
  \label{tab:nvs_interpolation}
\end{table}

\subsection{Content-Aware Frame Sampling}
\label{app:content_sampling}
Our default strategy selects context frames with stochastically perturbed temporal intervals and randomly samples intermediate frames as targets, using timestamps as a proxy for viewpoint change. Since a fixed temporal interval can correspond to very different baselines (\eg, a slow translational pan versus a fast rotation), we compare it against content-aware alternatives that use geometric cues to drive sampling:
\begin{itemize}
    \item \textbf{Context selection (A1).} We use rotation-compensated optical flow to estimate translational parallax and a coverage score to measure view overlap, selecting frames that balance geometric diversity with reliable shared content.
    \item \textbf{Target selection (B1).} We enforce an overlap ratio between $0.6$ and $0.9$ between each target view and the selected context views.
\end{itemize}
A0 and B0 denote our default temporal-jitter context sampling and random intermediate target sampling, respectively.

As shown in Table~\ref{tab:content_sampling}, flow-based context selection (A1B0) yields mixed changes on ScanNet, 7Scenes, and TUM, but clearer gains on MP3D (DAc/CD from 48.0/24.9 to 49.1/24.6). This suggests that the temporal heuristic is adequate for RE10K and DL3DV, whose camera trajectories are relatively regular, whereas content-aware selection is more beneficial under larger viewpoint variation. Overlap-aware target curation alone (A0B1) does not improve performance; its benefit emerges only when combined with flow-based context selection (A1B1), which yields the best MP3D result of 49.5/23.4. This indicates that moderate-overlap targets are most effective when the context views already provide sufficient parallax. Overall, temporal jitter offers a strong accuracy--simplicity trade-off on the datasets considered, while content-aware sampling is expected to become increasingly valuable as adaptation expands to more diverse videos with less regular camera motion.

\begin{table}[htbp]
  \centering
  \caption{\textbf{Temporal vs.\ content-aware frame sampling.}
  A0/B0: default temporal-jitter context sampling and random intermediate targets. A1: flow-based context selection. B1: overlap-curated targets.}
  \setlength{\tabcolsep}{3pt}
  \begin{tabular}{lcccccccc}
    \toprule
    \multirow{2}{*}{Method} & \multicolumn{2}{c}{ScanNet} & \multicolumn{2}{c}{7Scenes} & \multicolumn{2}{c}{TUM} & \multicolumn{2}{c}{MP3D} \\
    \cmidrule(lr){2-3} \cmidrule(lr){4-5} \cmidrule(lr){6-7} \cmidrule(lr){8-9}
    & DAc$\uparrow$ & CD$\downarrow$ & DAc$\uparrow$ & CD$\downarrow$ & DAc$\uparrow$ & CD$\downarrow$ & DAc$\uparrow$ & CD$\downarrow$ \\
    \midrule
    A0B0: temporal jitter (Ours) & 91.2 & 1.69 & \textbf{98.0} & \textbf{0.76} & 93.1 & \textbf{1.85} & 48.0 & 24.9 \\
    A1B0: flow-based context & \textbf{91.4} & \textbf{1.63} & 95.5 & 0.83 & \textbf{93.4} & 1.89 & 49.1 & 24.6 \\
    A0B1: curated targets & 90.6 & 1.70 & 96.4 & 0.89 & 92.3 & 1.86 & 48.0 & 25.1 \\
    A1B1: flow-based context + curated targets & 90.4 & \textbf{1.63} & 97.9 & 0.78 & 91.2 & 1.86 & \textbf{49.5} & \textbf{23.4} \\
    \bottomrule
  \end{tabular}
  \label{tab:content_sampling}
\end{table}

\subsection{Pose Estimation: Comparison with State-of-the-Art Foundation Models}
\label{app:pose_comparison}

In addition to the MV-DUSt3R~vs.~\name~comparison reported in the main paper (Table~\ref{tab:pose_estimation}), we further compare against three recent feed-forward foundation models that natively predict (or implicitly encode) camera parameters: $\pi^3$~\cite{wang2025pi}, Depth Anything 3 (DA3)~\cite{lin2025da3}, and VGGT~\cite{wang2025vggt}. Following the same RANSAC + PnP readout protocol as the main paper, we report the relative rotation error (RRE), relative translation error (RTE), and mean average error (mAE) on 7Scenes in Table~\ref{tab:pose_comparison_app}.

\begin{table}[h]
\centering
\small
\setlength{\tabcolsep}{8pt}
\caption{\textbf{Cross-method pose estimation on 7Scenes.}
\name~delivers competitive or superior pose accuracy over $\pi^3$, DA3, and VGGT, despite using only video-based weak supervision. Notably, VGGT+\name~surpasses $\pi^3$ on all three pose metrics.}
\label{tab:pose_comparison_app}
\begin{tabular}{lccc}
\toprule
\textbf{Method} & \textbf{RRE} $\downarrow$ & \textbf{RTE} $\downarrow$ & \textbf{mAE} $\downarrow$ \\
\midrule
$\pi^3$~\cite{wang2025pi}        & 0.3 & 3.2 & 15.4 \\
Depth Anything 3~\cite{lin2025da3} & 0.1 & 2.6 & 13.1 \\
VGGT~\cite{wang2025vggt}         & 0.4 & 2.7 & 12.9 \\
\midrule
\textbf{VGGT + \name}            & \textbf{0.1} & \textbf{2.1} & \textbf{12.1} \\
\bottomrule
\end{tabular}
\end{table}

This further demonstrates that the \name~recipe transfers consistently across architecturally distinct backbones (CroCo-based MV-DUSt3R and ViT-based VGGT), and that video-based adaptation can match or exceed pose-estimation quality of from-scratch foundation models trained with substantially larger 3D-supervised corpora.

\subsection{Robustness to Anchor Quality and Density}
\label{app:anchor_robustness}

A natural concern about \name's reliance on COLMAP-derived sparse anchors is brittleness: what happens when SfM points are sparse, or when they contain noise? Since our anchor loss is confidence-weighted, an adversarial concern is that the model could trivially down-weight all SfM points and thereby render the supervision inert. To rule out this failure mode, we conduct two controlled-perturbation experiments on the converged \name~(MV-DUSt3R, 10K) checkpoint.

\paragraph{Density: subsampling SfM anchors.}
We randomly retain a fraction of the COLMAP sparse points per training trajectory, holding all other settings fixed. Table~\ref{tab:anchor_density} reports Chamfer Distance ($\downarrow$) on the four reconstruction benchmarks under retentions of $100\%$ (default), $75\%$, $50\%$, and $25\%$.

\begin{table}[h]
\centering
\small
\setlength{\tabcolsep}{6pt}
\caption{\textbf{Anchor density ablation: Chamfer Distance ($\downarrow$) under random subsampling of SfM points.} \name~exhibits graceful degradation: even at $25\%$ retention, performance is largely preserved on three of four datasets, indicating that the dense differentiable consistency loss provides geometric supervision largely independent of anchor density.}
\label{tab:anchor_density}
\begin{tabular}{lcccc}
\toprule
\textbf{Retention} & \textbf{ScanNet} & \textbf{7Scenes} & \textbf{TUM} & \textbf{MP3D} \\
\midrule
100\% (default) & \textbf{1.69} & \textbf{0.76} & \textbf{1.85} & \textbf{24.9} \\
75\%            & 1.93          & 1.13          & 1.89          & 29.9 \\
50\%            & 1.98          & 1.31          & 1.89          & 28.5 \\
25\%            & 1.79          & 1.12          & 1.94          & 26.6 \\
\bottomrule
\end{tabular}
\end{table}

\paragraph{Quality: injecting Gaussian noise on SfM anchors.}
We perturb each retained 3D anchor point by isotropic Gaussian noise with standard deviation $\sigma = r \cdot \mathrm{median\_depth}$, where $r$ is the noise ratio. Table~\ref{tab:anchor_quality} reports Chamfer Distance ($\downarrow$) at $r \in \{5\%, 20\%, 50\%\}$.

\begin{table}[h]
\centering
\small
\setlength{\tabcolsep}{6pt}
\caption{\textbf{Anchor quality ablation: Chamfer Distance ($\downarrow$) under additive Gaussian noise on SfM anchor points.} Performance is essentially unchanged at $\sigma{=}5\%$ and degrades only mildly at $\sigma{=}20\%$. Only an aggressive $50\%$ perturbation causes severe drops, well beyond the noise levels expected from converged COLMAP reconstructions in practice.}
\label{tab:anchor_quality}
\begin{tabular}{lcccc}
\toprule
\textbf{Noise} $\sigma$ & \textbf{ScanNet} & \textbf{7Scenes} & \textbf{TUM} & \textbf{MP3D} \\
\midrule
5\%   & 1.92 & 0.96 & 1.85 & 27.2 \\
20\%  & 2.23 & 1.44 & 2.84 & 45.5 \\
50\%  & 3.15 & 5.03 & 7.28 & 68.3 \\
\bottomrule
\end{tabular}
\end{table}

\paragraph{Takeaway.}
Both experiments demonstrate \emph{graceful degradation}: \name~remains stable under aggressive sparsification (down to $25\%$ of anchors) and under realistic anchor noise ($\sigma{=}5\%$). This robustness stems from the complementary dense differentiable consistency loss (Sec.~\ref{sec:dense}), which provides geometric supervision orthogonal to anchor quality. The results also empirically rule out the failure mode where the confidence-weighted loss collapses all SfM contributions to zero: if it did, anchor perturbations would have no effect on performance, but Tables~\ref{tab:anchor_density} and~\ref{tab:anchor_quality} clearly show graded sensitivity to anchor quality.

\subsection{Novel View Synthesis Quality Comparison}

We evaluate the novel view synthesis performance of our method by rendering images from the predicted 3D Gaussians. Among the test datasets, we choose ScanNet~\cite{dai2017scannet} for this evaluation due to its high-quality imagery. In contrast, 7Scenes~\cite{shotton20137scenes} and TUM-RGBD~\cite{tum} often suffer from motion blur, and MP3D~\cite{chang2017matterport3d} lacks sufficient novel views with appropriate overlap, making them less suitable for consistent evaluation.
To ensure fairness, we allow training of the Gaussian head in our method, similar to the baseline. As shown in Table~\ref{tab:nvs_quality_scannet}, our method achieves slightly better perceptual quality than MV-DUSt3R in terms of PSNR, SSIM, and LPIPS. Although novel view synthesis is not our primary target, the improved rendering results further validate the quality of the reconstructed geometry.
\begin{table}[htbp]
  \centering
  % \small
  \setlength{\tabcolsep}{6pt}
  \caption{\textbf{Novel view synthesis quality on the ScanNet dataset.} 
  Comparison between MV-DUSt3R and our method (\name) using three standard metrics: PSNR ($\uparrow$), SSIM ($\uparrow$), and LPIPS ($\downarrow$). Higher PSNR/SSIM and lower LPIPS indicate better perceptual quality.}
  \begin{tabular}{cccc}
    \toprule
    \textbf{Method} & \textbf{PSNR}$\uparrow$ & \textbf{SSIM}$\uparrow$ & \textbf{LPIPS}$\downarrow$ \\
    \midrule
    MV-DUSt3R~\cite{tang2024mvdust3r} & 20.72 & 0.66 & 0.188 \\
    \name~(Ours)                     & \textbf{20.98} & \textbf{0.67} & \textbf{0.184} \\
    \bottomrule
  \end{tabular}
  \label{tab:nvs_quality_scannet}
\end{table}

\subsection{The Impact of Scene Diversity vs. Sample Density}
To investigate whether the generalization capability stems from the total volume of training samples or the diversity of the scenes, we conduct a controlled scaling experiment. We fix the total number of training iterations and the total sample budget to ensure a fair comparison. We then vary the number of unique training video scenes, ranging from 100 to 2,000. In this setup, as the number of scenes increases, the number of samples drawn from each individual video decreases proportionally. We evaluate the zero-shot generalization performance on the 7-Scenes and TUM datasets using the Chamfer Distance (CD) as the primary metric.

\begin{figure}[ht]
\centering
\includegraphics[width=0.7\textwidth]{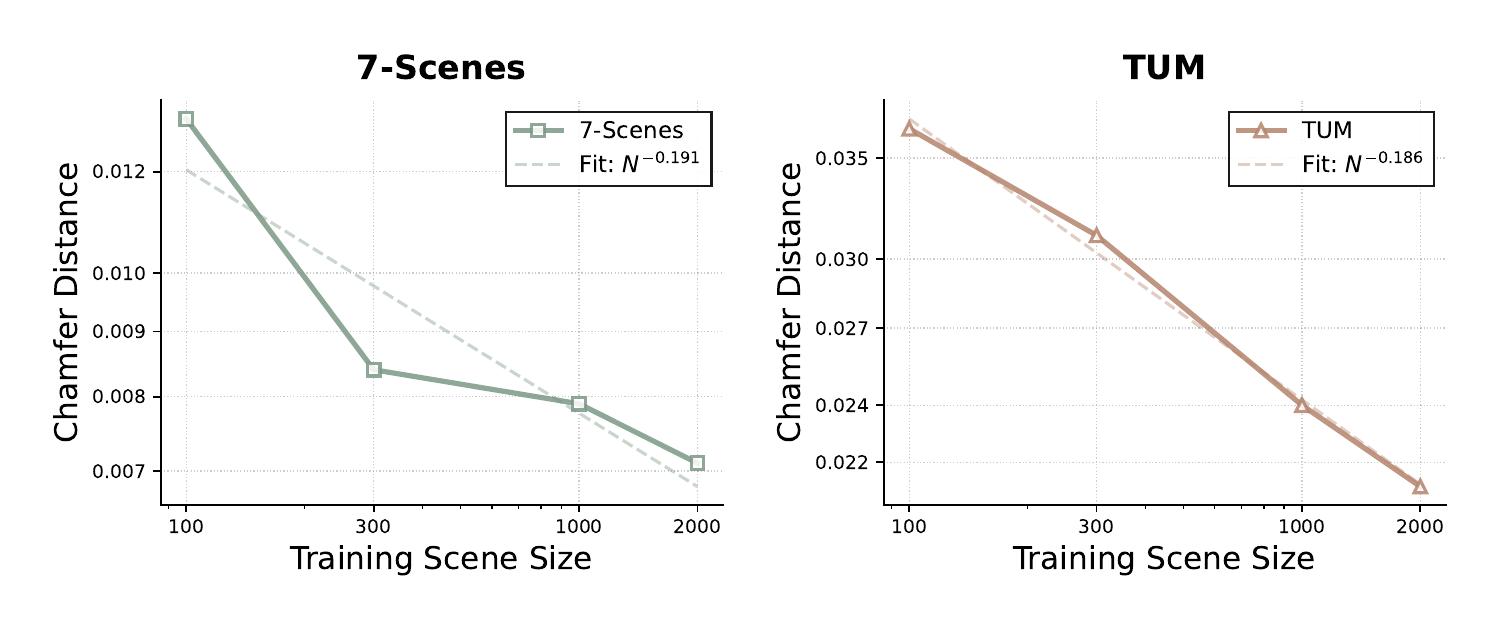}
\caption{
Generalization performance under a fixed number of training iterations, as the diversity of training scenes increases from 100 to 2K video scenes, demonstrating consistent performance gains from broader scene coverage.
}
\label{fig:scene}
\end{figure}

The results, illustrated in Figure~\ref{fig:scene}, reveal a clear power-law relationship between training scene diversity and generalization error. On both the 7-Scenes and TUM datasets, we observe a consistent and significant reduction in Chamfer Distance as the number of training scenes increases. These findings underscore the necessity of leveraging large-scale, diverse internet video data. The diminishing returns of over-sampling from the same scene suggest that the model benefits more from exposure to novel geometric contexts and environmental variations, which are abundant in internet-scale datasets.

Crucially, since the total number of training iterations is kept constant, this performance gain is achieved by training on more diverse scenes with fewer samples per scene, rather than seeing more frames from a limited set of environments.  This provides empirical evidence that scene-level diversity (the "breadth" of data) is a more critical factor for robust generalization than the temporal sampling density within a single video (the "depth" of data).

% \subsection{Experiments of Depth any video}
\section{Limitation and Impact}
\label{App_limit}
% \vspace{-4mm}
\paragraph{Limitation and Further Discussion.} 
% Our method builds on pretrained 3D reconstruction model designed for static scenes and relies on Internet video sequences that predominantly depict static environments. We carefully filter out clips with excessive motion to ensure reliable sparse 3D supervision. 
% However, moving objects in training videos can still introduce inconsistent geometry and noisy supervision, limiting the applicability to dynamic scenes. 
Our method builds on a pretrained sparse 3D reconstruction model for static scenes and we primarily leverage Internet video sequences that capture largely static environments. Although we filter out clips with excessive motion to ensure reliable sparse 3D supervision, this strategy cannot fully exploit the vast scale and diversity of Internet video data, where dynamic content is prevalent. Extending our framework to pretrained models capable of handling both static and dynamic scenes represents a highly promising direction, with the potential to unlock the full value of large-scale Internet videos, enhance model generalization across diverse environments, and alleviate the longstanding scarcity and limited diversity of existing 3D training datasets.
\paragraph{Impact Statement}
This work presents a method for adapting 3D geometric foundation models using weak supervision from internet videos. The proposed approach has the potential to significantly enhance the capabilities of 3D vision systems, enabling them to better understand and interpret complex real-world environments. This advancement could lead to improved applications in various fields, including robotics, augmented reality, and autonomous driving.

\section{Qualitative Results}
\label{app_vis}
\subsection{Qualitative 3D Reconstruction Results}
\begin{figure*}[ht]
\centering
\includegraphics[width = 1.0\textwidth]{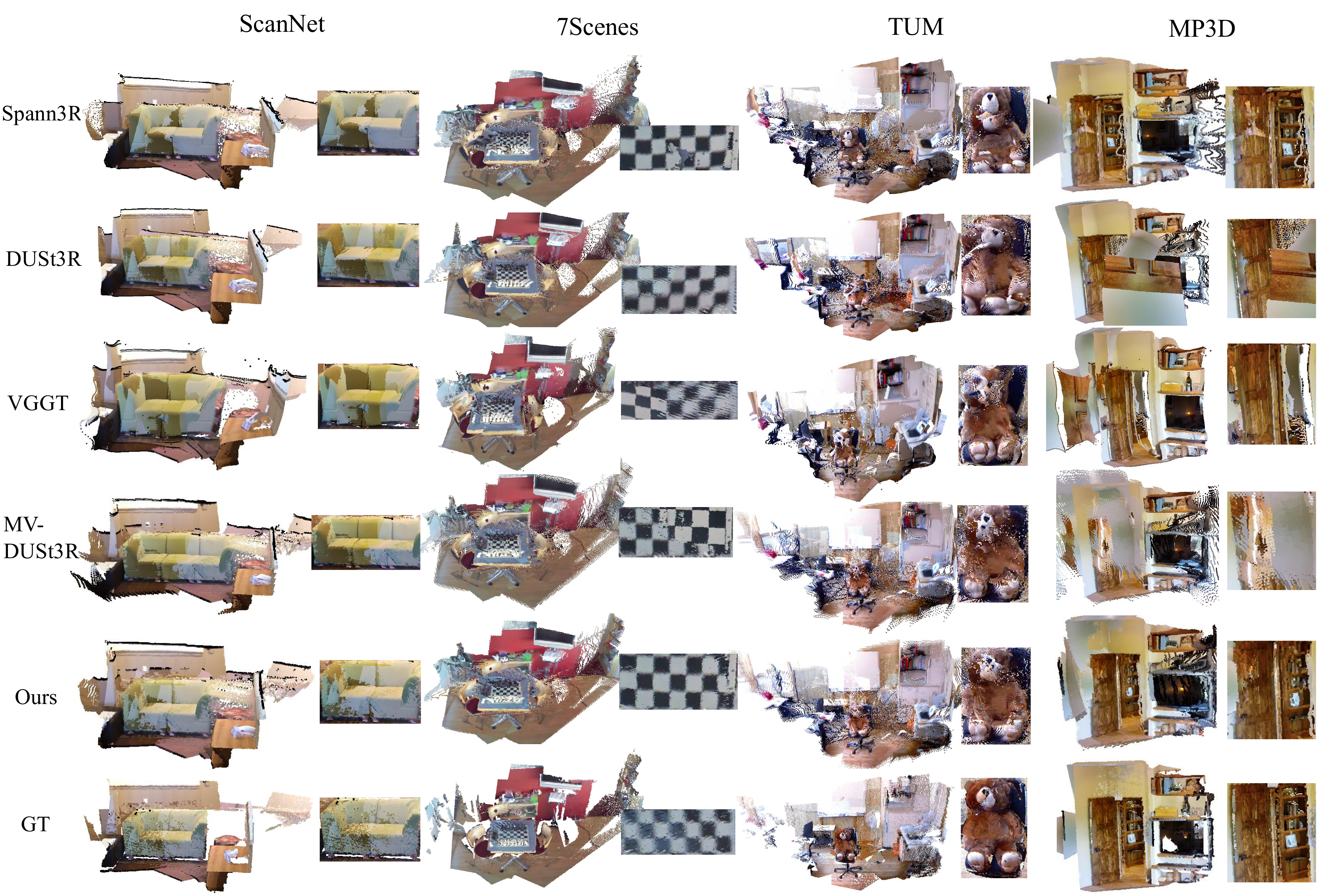}
  \caption{Qualitative comparison of reconstructed point clouds across different methods, alongside ground-truth geometry. Zoomed-in views of representative regions are shown on the right. (Note: The 7Scenes ground-truth may appear slightly misaligned due to minor pose inaccuracies in the dataset.)}
  \label{fig:vis_comp}
\end{figure*}

Qualitative reconstruction comparisons against MV-DUSt3R and VGGT
in Fig~\ref{fig:vis_comp} show that SAGE yields more geometrically consistent reconstructions on its respective backbones.

\subsection{Additional Qualitative 3D Reconstruction Results}
\begin{figure*}[ht]
\centering
\includegraphics[width = 1.0\textwidth]{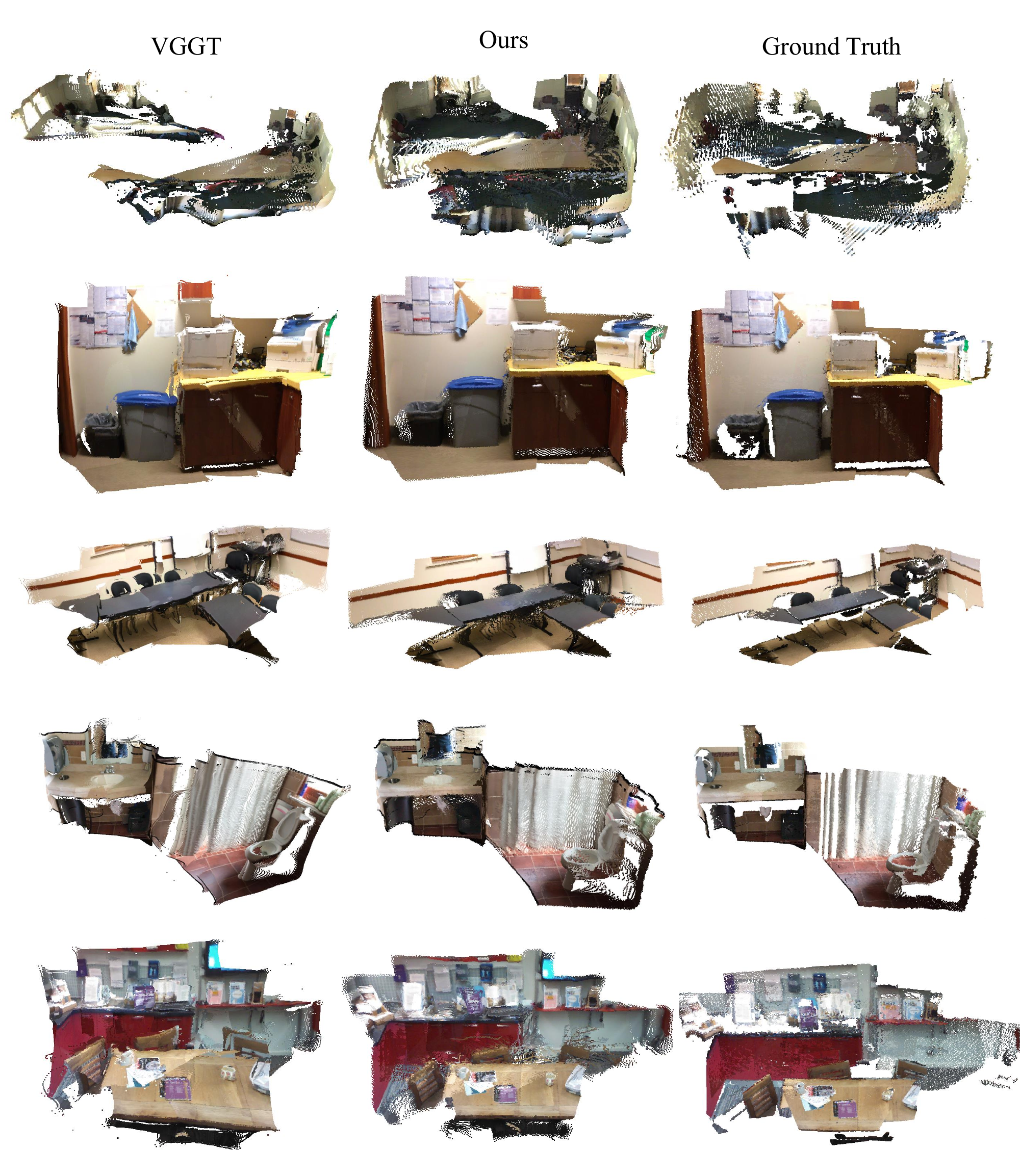}
  \caption{Qualitative results of reconstructed point clouds with 8 input views across diverse scenes, shown alongside ground-truth geometry.
All examples are drawn from unseen scenes and demonstrate the generalization ability of the model.}
  \label{fig:supply_vis}
\end{figure*}

\begin{figure*}[ht]
\centering
\includegraphics[width = 1.0\textwidth]{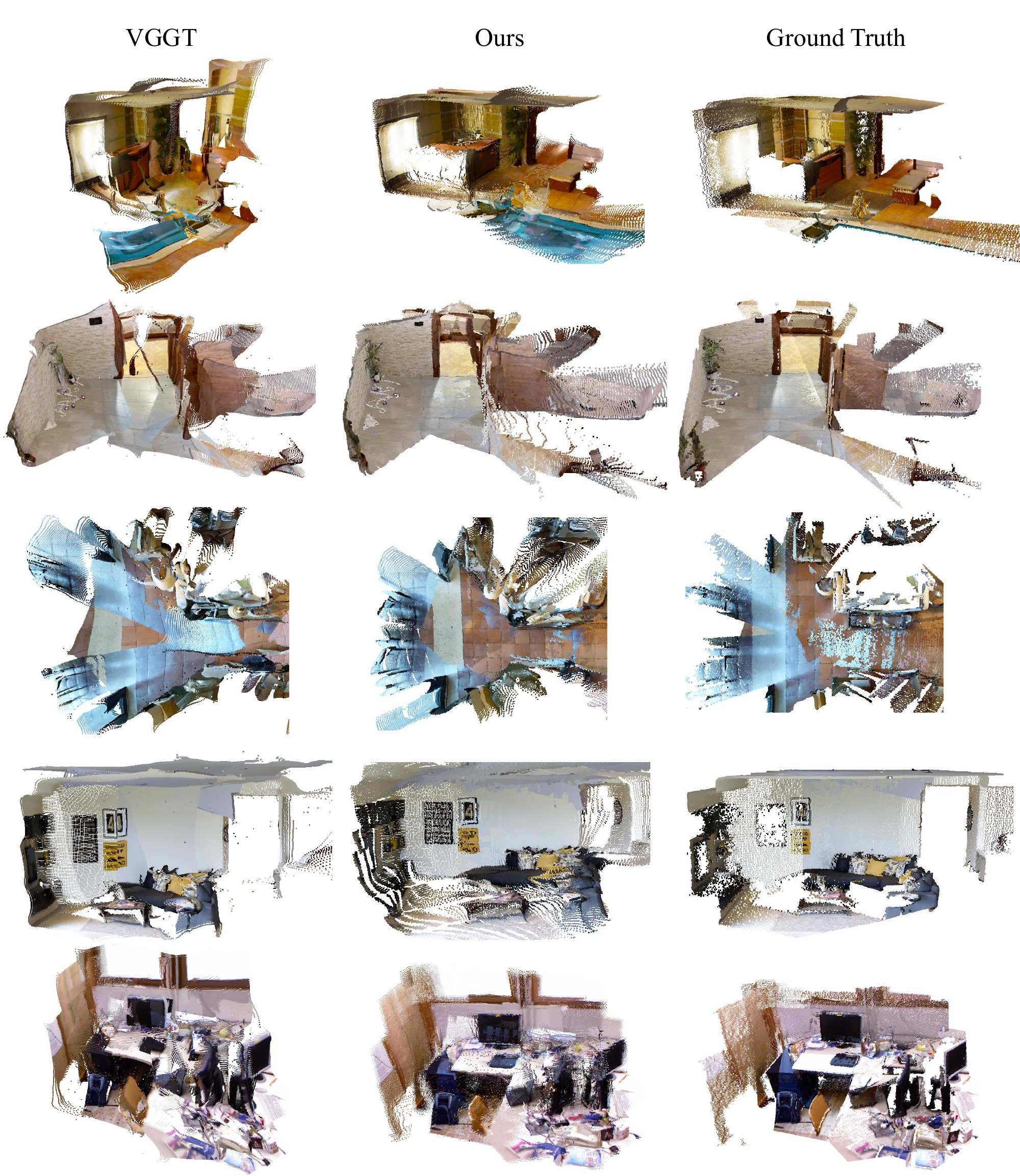}
  \caption{Qualitative comparisons results of reconstructed point clouds views on challenging scenes, shown alongside ground-truth geometry.
These examples feature low image overlap, texture-less surfaces (e.g., walls or floors), or inherently difficult matching conditions, and are drawn from previously unseen test scenes.}
  \label{fig:supply_vis2}
\end{figure*}

We provide additional qualitative reconstruction results in Figure~\ref{fig:supply_vis}, and further examples on more challenging scenes, low overlap or limited texture, in Figure~\ref{fig:supply_vis2}.

\subsection{Qualitative Rendering Results}
\label{Render_vis}
\begin{figure*}[ht]
\centering
\includegraphics[width = 1.0\textwidth]{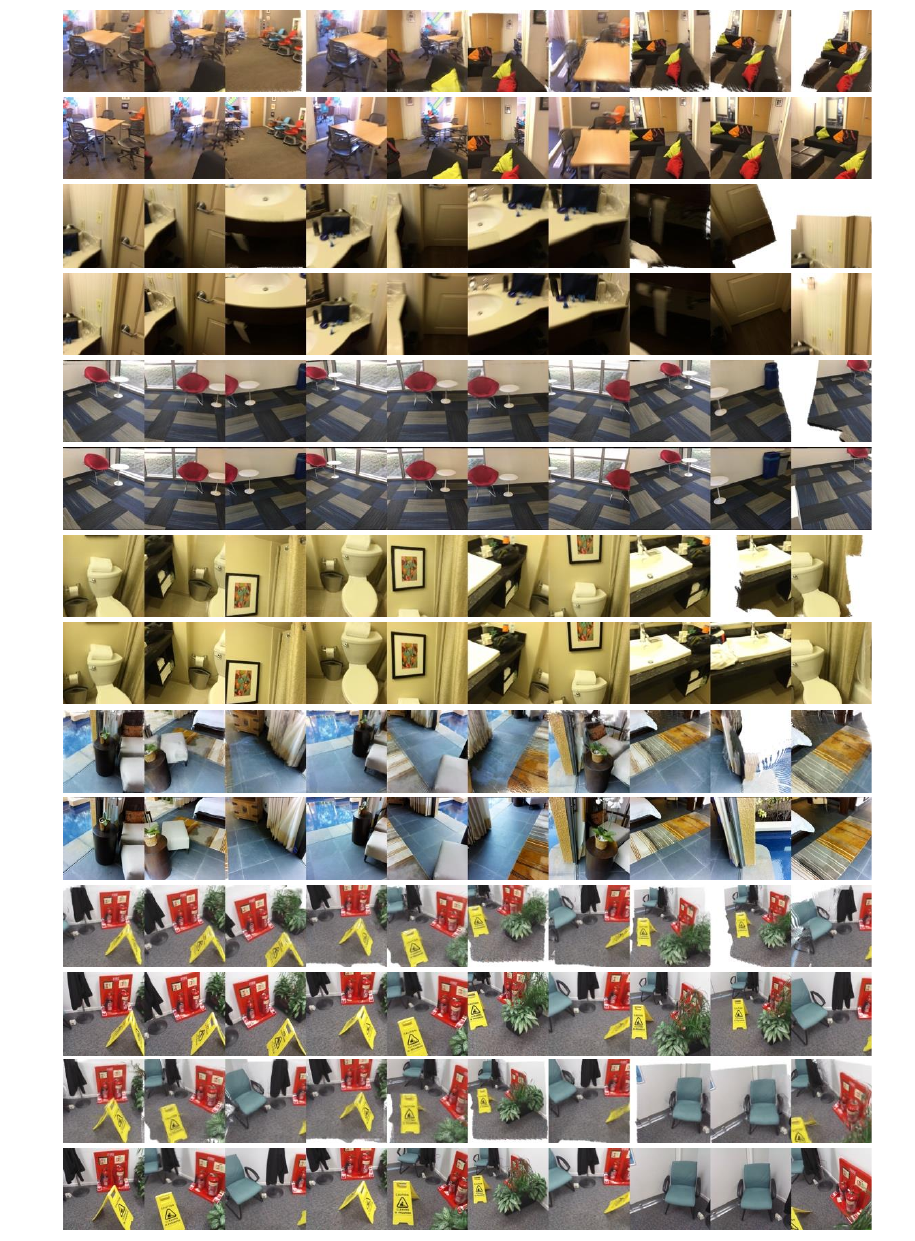}
\caption{Qualitative visualization of rendered images and ground-truth views.}
\label{fig:render_vis}
\end{figure*}

\begin{figure*}[ht]
\centering
\includegraphics[width = 1.0\textwidth]{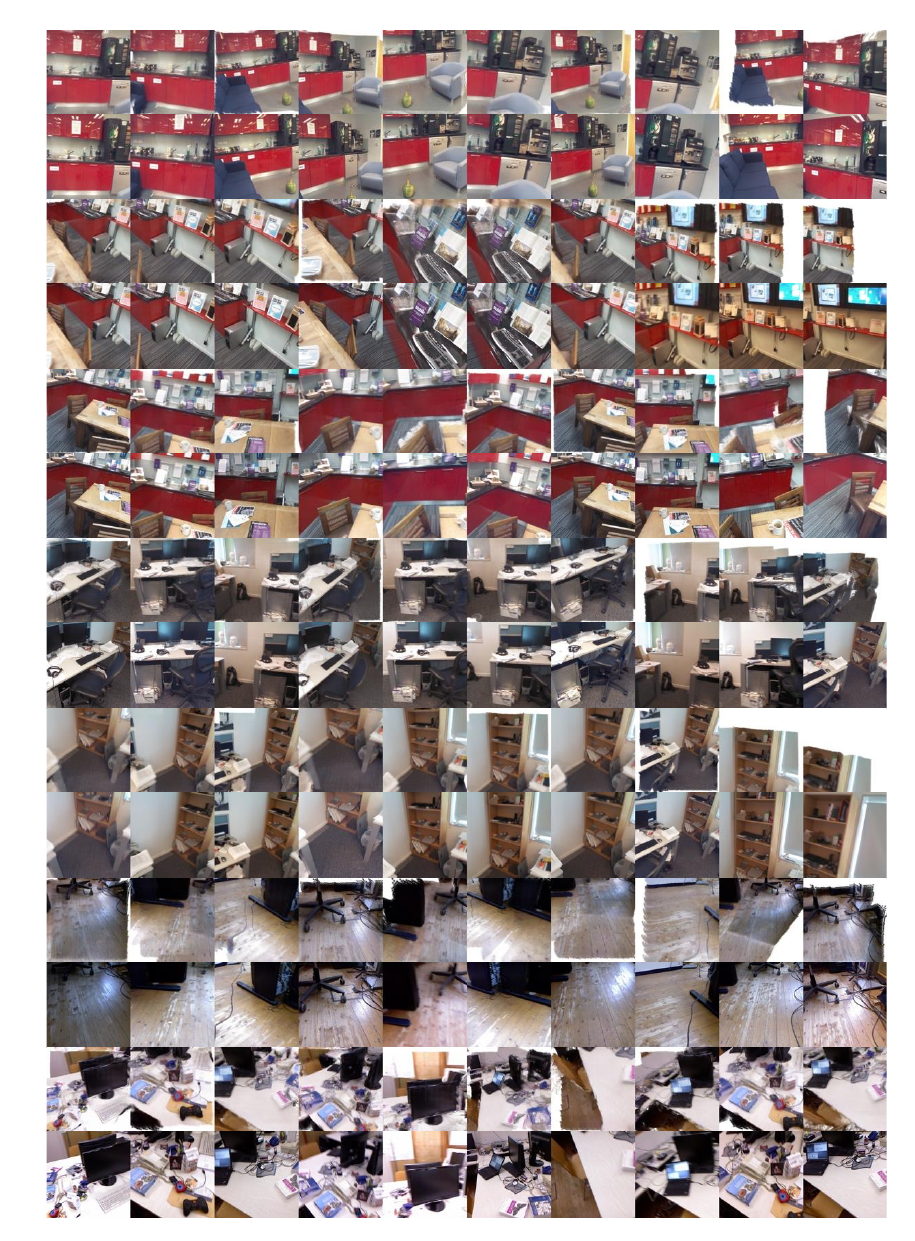}
  \caption{Qualitative visualization of rendered images and ground-truth views. 8 images are provided as model input, from which a 3D Gaussian representation is inferred and rendered to reconstruct the original 8 views as well as 2 novel viewpoints. Due to occlusions, novel-view renderings are limited to regions covered by the input views. The rendered images exhibit high visual fidelity and consistency with ground truth.}
  \label{fig:render_vis2}
\end{figure*}

We visualize the rendering quality of the inferred 3D representation in Figure~\ref{fig:render_vis} and Figure~\ref{fig:render_vis2}. Given 8 input images as model input, the network performs a forward pass to infer a 3D Gaussian representation, which is then used to render images from both seen viewpoints and novel viewpoints. Specifically, the reconstructed representation is rendered to reproduce the original 8 input views, as well as 2 additional novel views for evaluating generalization to unseen viewpoints.
Due to occlusions and limited field-of-view coverage in the input images, the novel-view renderings are naturally restricted to regions observed by the input views. Despite this limitation, the rendered images exhibit high visual fidelity, sharp geometric structures, and strong consistency with the ground-truth images, demonstrating that the learned 3D Gaussian representation effectively captures the underlying scene geometry and appearance. These qualitative results indicate that the proposed model can reconstruct accurate 3D scene representations from sparse multi-view inputs and generate high-quality renderings across viewpoints.

\clearpage
\newpage
\section*{NeurIPS Paper Checklist}

\begin{enumerate}

\item {\bf Claims}
    \item[] Question: Do the main claims made in the abstract and introduction accurately reflect the paper's contributions and scope?
    \item[] Answer: \answerYes{} % Replace by \answerYes{}, \answerNo{}, or \answerNA{}.
    \item[] Justification: We state our claims clearly in the abstract and introduction. The claims are supported by various experiments.
    \item[] Guidelines:
    \begin{itemize}
        \item The answer \answerNA{} means that the abstract and introduction do not include the claims made in the paper.
        \item The abstract and/or introduction should clearly state the claims made, including the contributions made in the paper and important assumptions and limitations. A \answerNo{} or \answerNA{} answer to this question will not be perceived well by the reviewers. 
        \item The claims made should match theoretical and experimental results, and reflect how much the results can be expected to generalize to other settings. 
        \item It is fine to include aspirational goals as motivation as long as it is clear that these goals are not attained by the paper. 
    \end{itemize}

\item {\bf Limitations}
    \item[] Question: Does the paper discuss the limitations of the work performed by the authors?
    \item[] Answer: \answerYes{} % Replace by \answerYes{}, \answerNo{}, or \answerNA{}.
    \item[] Justification: We discuss the limitations in App~\ref{App_limit}.
    \item[] Guidelines: 
    \begin{itemize}
        \item The answer \answerNA{} means that the paper has no limitation while the answer \answerNo{} means that the paper has limitations, but those are not discussed in the paper. 
        \item The authors are encouraged to create a separate ``Limitations'' section in their paper.
        \item The paper should point out any strong assumptions and how robust the results are to violations of these assumptions (e.g., independence assumptions, noiseless settings, model well-specification, asymptotic approximations only holding locally). The authors should reflect on how these assumptions might be violated in practice and what the implications would be.
        \item The authors should reflect on the scope of the claims made, e.g., if the approach was only tested on a few datasets or with a few runs. In general, empirical results often depend on implicit assumptions, which should be articulated.
        \item The authors should reflect on the factors that influence the performance of the approach. For example, a facial recognition algorithm may perform poorly when image resolution is low or images are taken in low lighting. Or a speech-to-text system might not be used reliably to provide closed captions for online lectures because it fails to handle technical jargon.
        \item The authors should discuss the computational efficiency of the proposed algorithms and how they scale with dataset size.
        \item If applicable, the authors should discuss possible limitations of their approach to address problems of privacy and fairness.
        \item While the authors might fear that complete honesty about limitations might be used by reviewers as grounds for rejection, a worse outcome might be that reviewers discover limitations that aren't acknowledged in the paper. The authors should use their best judgment and recognize that individual actions in favor of transparency play an important role in developing norms that preserve the integrity of the community. Reviewers will be specifically instructed to not penalize honesty concerning limitations.
    \end{itemize}

\item {\bf Theory assumptions and proofs}
    \item[] Question: For each theoretical result, does the paper provide the full set of assumptions and a complete (and correct) proof?
    \item[] Answer: \answerNA{} % Replace by \answerYes{}, \answerNo{}, or \answerNA{}.
    \item[] Justification: The paper does not have any theoretical result
    \item[] Guidelines:
    \begin{itemize}
        \item The answer \answerNA{} means that the paper does not include theoretical results. 
        \item All the theorems, formulas, and proofs in the paper should be numbered and cross-referenced.
        \item All assumptions should be clearly stated or referenced in the statement of any theorems.
        \item The proofs can either appear in the main paper or the supplemental material, but if they appear in the supplemental material, the authors are encouraged to provide a short proof sketch to provide intuition. 
        \item Inversely, any informal proof provided in the core of the paper should be complemented by formal proofs provided in appendix or supplemental material.
        \item Theorems and Lemmas that the proof relies upon should be properly referenced. 
    \end{itemize}

    \item {\bf Experimental result reproducibility}
    \item[] Question: Does the paper fully disclose all the information needed to reproduce the main experimental results of the paper to the extent that it affects the main claims and/or conclusions of the paper (regardless of whether the code and data are provided or not)?
    \item[] Answer: \answerYes{} % Replace by \answerYes{}, \answerNo{}, or \answerNA{}.
    \item[] Justification: We provide the complete experimental setup and data in the App.~\ref{app_details}, ensuring the reproducibility of the results.
    \item[] Guidelines:
    \begin{itemize}
        \item The answer \answerNA{} means that the paper does not include experiments.
        \item If the paper includes experiments, a \answerNo{} answer to this question will not be perceived well by the reviewers: Making the paper reproducible is important, regardless of whether the code and data are provided or not.
        \item If the contribution is a dataset and\slash or model, the authors should describe the steps taken to make their results reproducible or verifiable. 
        \item Depending on the contribution, reproducibility can be accomplished in various ways. For example, if the contribution is a novel architecture, describing the architecture fully might suffice, or if the contribution is a specific model and empirical evaluation, it may be necessary to either make it possible for others to replicate the model with the same dataset, or provide access to the model. In general. releasing code and data is often one good way to accomplish this, but reproducibility can also be provided via detailed instructions for how to replicate the results, access to a hosted model (e.g., in the case of a large language model), releasing of a model checkpoint, or other means that are appropriate to the research performed.
        \item While NeurIPS does not require releasing code, the conference does require all submissions to provide some reasonable avenue for reproducibility, which may depend on the nature of the contribution. For example
        \begin{enumerate}
            \item If the contribution is primarily a new algorithm, the paper should make it clear how to reproduce that algorithm.
            \item If the contribution is primarily a new model architecture, the paper should describe the architecture clearly and fully.
            \item If the contribution is a new model (e.g., a large language model), then there should either be a way to access this model for reproducing the results or a way to reproduce the model (e.g., with an open-source dataset or instructions for how to construct the dataset).
            \item We recognize that reproducibility may be tricky in some cases, in which case authors are welcome to describe the particular way they provide for reproducibility. In the case of closed-source models, it may be that access to the model is limited in some way (e.g., to registered users), but it should be possible for other researchers to have some path to reproducing or verifying the results.
        \end{enumerate}
    \end{itemize}

\item {\bf Open access to data and code}
    \item[] Question: Does the paper provide open access to the data and code, with sufficient instructions to faithfully reproduce the main experimental results, as described in supplemental material?
    \item[] Answer: \answerYes{} % Replace by \answerYes{}, \answerNo{}, or \answerNA{}.
    \item[] Justification: Code is publicly available at \url{https://github.com/aim-uofa/SAGE}. The training videos come from the public RealEstate10K and DL3DV datasets, and the data processing pipeline and training details are described in App.~\ref{app:data_pocess} and App.~\ref{app_details}.
    \item[] Guidelines:
    \begin{itemize}
        \item The answer \answerNA{} means that paper does not include experiments requiring code.
        \item Please see the NeurIPS code and data submission guidelines (\url{https://neurips.cc/public/guides/CodeSubmissionPolicy}) for more details.
        \item While we encourage the release of code and data, we understand that this might not be possible, so \answerNo{} is an acceptable answer. Papers cannot be rejected simply for not including code, unless this is central to the contribution (e.g., for a new open-source benchmark).
        \item The instructions should contain the exact command and environment needed to run to reproduce the results. See the NeurIPS code and data submission guidelines (\url{https://neurips.cc/public/guides/CodeSubmissionPolicy}) for more details.
        \item The authors should provide instructions on data access and preparation, including how to access the raw data, preprocessed data, intermediate data, and generated data, etc.
        \item The authors should provide scripts to reproduce all experimental results for the new proposed method and baselines. If only a subset of experiments are reproducible, they should state which ones are omitted from the script and why.
        \item At submission time, to preserve anonymity, the authors should release anonymized versions (if applicable).
        \item Providing as much information as possible in supplemental material (appended to the paper) is recommended, but including URLs to data and code is permitted.
    \end{itemize}

\item {\bf Experimental setting/details}
    \item[] Question: Does the paper specify all the training and test details (e.g., data splits, hyperparameters, how they were chosen, type of optimizer) necessary to understand the results?
    \item[] Answer: \answerYes{} % Replace by \answerYes{}, \answerNo{}, or \answerNA{}.
    \item[] Justification: We discussed the experimental setup and hyperparameter selection in detail in the appendix to ensure the reproducibility of the results.
    \item[] Guidelines:
    \begin{itemize}
        \item The answer \answerNA{} means that the paper does not include experiments.
        \item The experimental setting should be presented in the core of the paper to a level of detail that is necessary to appreciate the results and make sense of them.
        \item The full details can be provided either with the code, in appendix, or as supplemental material.
    \end{itemize}

\item {\bf Experiment statistical significance}
    \item[] Question: Does the paper report error bars suitably and correctly defined or other appropriate information about the statistical significance of the experiments?
    \item[] Answer: \answerNo{} % Replace by \answerYes{}, \answerNo{}, or \answerNA{}.
    \item[] Justification: We did not report error bars or confidence intervals in the paper.
    \item[] Guidelines:
    \begin{itemize}
        \item The answer \answerNA{} means that the paper does not include experiments.
        \item The authors should answer \answerYes{} if the results are accompanied by error bars, confidence intervals, or statistical significance tests, at least for the experiments that support the main claims of the paper.
        \item The factors of variability that the error bars are capturing should be clearly stated (for example, train/test split, initialization, random drawing of some parameter, or overall run with given experimental conditions).
        \item The method for calculating the error bars should be explained (closed form formula, call to a library function, bootstrap, etc.)
        \item The assumptions made should be given (e.g., Normally distributed errors).
        \item It should be clear whether the error bar is the standard deviation or the standard error of the mean.
        \item It is OK to report 1-sigma error bars, but one should state it. The authors should preferably report a 2-sigma error bar than state that they have a 96\% CI, if the hypothesis of Normality of errors is not verified.
        \item For asymmetric distributions, the authors should be careful not to show in tables or figures symmetric error bars that would yield results that are out of range (e.g., negative error rates).
        \item If error bars are reported in tables or plots, the authors should explain in the text how they were calculated and reference the corresponding figures or tables in the text.
    \end{itemize}

\item {\bf Experiments compute resources}
    \item[] Question: For each experiment, does the paper provide sufficient information on the computer resources (type of compute workers, memory, time of execution) needed to reproduce the experiments?
    \item[] Answer: \answerYes{} % Replace by \answerYes{}, \answerNo{}, or \answerNA{}.
    \item[] Justification: We provided the amount of compute required in experimental details.
    \item[] Guidelines:
    \begin{itemize}
        \item The answer \answerNA{} means that the paper does not include experiments.
        \item The paper should indicate the type of compute workers CPU or GPU, internal cluster, or cloud provider, including relevant memory and storage.
        \item The paper should provide the amount of compute required for each of the individual experimental runs as well as estimate the total compute. 
        \item The paper should disclose whether the full research project required more compute than the experiments reported in the paper (e.g., preliminary or failed experiments that didn't make it into the paper). 
    \end{itemize}
    
\item {\bf Code of ethics}
    \item[] Question: Does the research conducted in the paper conform, in every respect, with the NeurIPS Code of Ethics \url{https://neurips.cc/public/EthicsGuidelines}?
    \item[] Answer: \answerYes{} % Replace by \answerYes{}, \answerNo{}, or \answerNA{}.
    \item[] Justification: We do not foresee any such issue in our work
    \item[] Guidelines:
    \begin{itemize}
        \item The answer \answerNA{} means that the authors have not reviewed the NeurIPS Code of Ethics.
        \item If the authors answer \answerNo, they should explain the special circumstances that require a deviation from the Code of Ethics.
        \item The authors should make sure to preserve anonymity (e.g., if there is a special consideration due to laws or regulations in their jurisdiction).
    \end{itemize}

\item {\bf Broader impacts}
    \item[] Question: Does the paper discuss both potential positive societal impacts and negative societal impacts of the work performed?
    \item[] Answer: \answerYes{} % Replace by \answerYes{}, \answerNo{}, or \answerNA{}.
    \item[] Justification: We discussed the potential social impacts in App.~\ref{App_limit}.
    \item[] Guidelines:
    \begin{itemize}
        \item The answer \answerNA{} means that there is no societal impact of the work performed.
        \item If the authors answer \answerNA{} or \answerNo, they should explain why their work has no societal impact or why the paper does not address societal impact.
        \item Examples of negative societal impacts include potential malicious or unintended uses (e.g., disinformation, generating fake profiles, surveillance), fairness considerations (e.g., deployment of technologies that could make decisions that unfairly impact specific groups), privacy considerations, and security considerations.
        \item The conference expects that many papers will be foundational research and not tied to particular applications, let alone deployments. However, if there is a direct path to any negative applications, the authors should point it out. For example, it is legitimate to point out that an improvement in the quality of generative models could be used to generate Deepfakes for disinformation. On the other hand, it is not needed to point out that a generic algorithm for optimizing neural networks could enable people to train models that generate Deepfakes faster.
        \item The authors should consider possible harms that could arise when the technology is being used as intended and functioning correctly, harms that could arise when the technology is being used as intended but gives incorrect results, and harms following from (intentional or unintentional) misuse of the technology.
        \item If there are negative societal impacts, the authors could also discuss possible mitigation strategies (e.g., gated release of models, providing defenses in addition to attacks, mechanisms for monitoring misuse, mechanisms to monitor how a system learns from feedback over time, improving the efficiency and accessibility of ML).
    \end{itemize}
    
\item {\bf Safeguards}
    \item[] Question: Does the paper describe safeguards that have been put in place for responsible release of data or models that have a high risk for misuse (e.g., pre-trained language models, image generators, or scraped datasets)?
    \item[] Answer: \answerNA{} % Replace by \answerYes{}, \answerNo{}, or \answerNA{}.
    \item[] Justification:  Not applicable
    \item[] Guidelines:
    \begin{itemize}
        \item The answer \answerNA{} means that the paper poses no such risks.
        \item Released models that have a high risk for misuse or dual-use should be released with necessary safeguards to allow for controlled use of the model, for example by requiring that users adhere to usage guidelines or restrictions to access the model or implementing safety filters. 
        \item Datasets that have been scraped from the Internet could pose safety risks. The authors should describe how they avoided releasing unsafe images.
        \item We recognize that providing effective safeguards is challenging, and many papers do not require this, but we encourage authors to take this into account and make a best faith effort.
    \end{itemize}

\item {\bf Licenses for existing assets}
    \item[] Question: Are the creators or original owners of assets (e.g., code, data, models), used in the paper, properly credited and are the license and terms of use explicitly mentioned and properly respected?
    \item[] Answer: \answerYes{} % Replace by \answerYes{}, \answerNo{}, or \answerNA{}.
    \item[] Justification:  We cited the original papers that produced the code packages and datasets used in this paper.
    \item[] Guidelines:
    \begin{itemize}
        \item The answer \answerNA{} means that the paper does not use existing assets.
        \item The authors should cite the original paper that produced the code package or dataset.
        \item The authors should state which version of the asset is used and, if possible, include a URL.
        \item The name of the license (e.g., CC-BY 4.0) should be included for each asset.
        \item For scraped data from a particular source (e.g., website), the copyright and terms of service of that source should be provided.
        \item If assets are released, the license, copyright information, and terms of use in the package should be provided. For popular datasets, \url{paperswithcode.com/datasets} has curated licenses for some datasets. Their licensing guide can help determine the license of a dataset.
        \item For existing datasets that are re-packaged, both the original license and the license of the derived asset (if it has changed) should be provided.
        \item If this information is not available online, the authors are encouraged to reach out to the asset's creators.
    \end{itemize}

\item {\bf New assets}
    \item[] Question: Are new assets introduced in the paper well documented and is the documentation provided alongside the assets?
    \item[] Answer: \answerYes{} % Replace by \answerYes{}, \answerNo{}, or \answerNA{}.
    \item[] Justification: We provide the details in the App.~\ref{app:data_pocess}.
    \item[] Guidelines:
    \begin{itemize}
        \item The answer \answerNA{} means that the paper does not release new assets.
        \item Researchers should communicate the details of the dataset\slash code\slash model as part of their submissions via structured templates. This includes details about training, license, limitations, etc. 
        \item The paper should discuss whether and how consent was obtained from people whose asset is used.
        \item At submission time, remember to anonymize your assets (if applicable). You can either create an anonymized URL or include an anonymized zip file.
    \end{itemize}

\item {\bf Crowdsourcing and research with human subjects}
    \item[] Question: For crowdsourcing experiments and research with human subjects, does the paper include the full text of instructions given to participants and screenshots, if applicable, as well as details about compensation (if any)? 
    \item[] Answer: \answerNA{} % Replace by \answerYes{}, \answerNo{}, or \answerNA{}.
    \item[] Justification: Not applicable.
    \item[] Guidelines:
    \begin{itemize}
        \item The answer \answerNA{} means that the paper does not involve crowdsourcing nor research with human subjects.
        \item Including this information in the supplemental material is fine, but if the main contribution of the paper involves human subjects, then as much detail as possible should be included in the main paper. 
        \item According to the NeurIPS Code of Ethics, workers involved in data collection, curation, or other labor should be paid at least the minimum wage in the country of the data collector. 
    \end{itemize}

\item {\bf Institutional review board (IRB) approvals or equivalent for research with human subjects}
    \item[] Question: Does the paper describe potential risks incurred by study participants, whether such risks were disclosed to the subjects, and whether Institutional Review Board (IRB) approvals (or an equivalent approval/review based on the requirements of your country or institution) were obtained?
    \item[] Answer: \answerNA{} % Replace by \answerYes{}, \answerNo{}, or \answerNA{}.
    \item[] Justification: Not applicable.
    \item[] Guidelines:
    \begin{itemize}
        \item The answer \answerNA{} means that the paper does not involve crowdsourcing nor research with human subjects.
        \item Depending on the country in which research is conducted, IRB approval (or equivalent) may be required for any human subjects research. If you obtained IRB approval, you should clearly state this in the paper. 
        \item We recognize that the procedures for this may vary significantly between institutions and locations, and we expect authors to adhere to the NeurIPS Code of Ethics and the guidelines for their institution. 
        \item For initial submissions, do not include any information that would break anonymity (if applicable), such as the institution conducting the review.
    \end{itemize}

\item {\bf Declaration of LLM usage}
    \item[] Question: Does the paper describe the usage of LLMs if it is an important, original, or non-standard component of the core methods in this research? Note that if the LLM is used only for writing, editing, or formatting purposes and does \emph{not} impact the core methodology, scientific rigor, or originality of the research, declaration is not required.
    %this research? 
    \item[] Answer: \answerNA{}. % Replace by \answerYes{}, \answerNo{}, or \answerNA{}.
    \item[] Justification: This is not applicable to our research.
    \item[] Guidelines:
    \begin{itemize}
        \item The answer \answerNA{} means that the core method development in this research does not involve LLMs as any important, original, or non-standard components.
        \item Please refer to our LLM policy in the NeurIPS handbook for what should or should not be described.
    \end{itemize}

\end{enumerate}

\end{document}